# Pruning Attention Heads of Transformer Models Using *A** Search

A NOVEL APPROACH TO COMPRESS BIG NLP ARCHITECTURES

Archit Parnami*[1], Rahul Singh[2], Tarun Joshi[2]
[1]*University of North Carolina at Charlotte,* [2]*Corporate Model Risk, Wells Fargo, USA*

## Abstract

Recent years have seen a growing adoption of Transformer models such as BERT in Natural Language Processing and even in Computer Vision. However, due to their size, there has been limited adoption of such models within resource-constrained computing environments This paper proposes novel pruning algorithm to compress transformer models by eliminating redundant Attention Heads. We apply the $A^*$ search algorithm to obtain a pruned model with strict accuracy guarantees. Our results indicate that the method could eliminate as much as 40% of the attention heads in the BERT transformer model with no loss in accuracy.

## 1 Introduction

Language models such as BERT [1], GPT-2 [2], GPT-3 [3], and many others that are based on transformer architectures have pushed the boundaries on many Natural Language Processing (NLP) tasks. These models are initially trained on big datasets like C4 [1], Wikipedia [2] and fine-tuned on downstream tasks like Natural Language Translation, Document Summarization, Document Generation, Named Entity Recognition, and Video Understanding. This methodology is also called Transfer Learning. Although there has been considerable performance gain, the usage of transformer-based models comes at the cost of huge memory size (see Figure 1). For example, BERT-large has 340 million parameters and the most recent GPT-3 has 175 billion parameters [1, 3]. Their size causes several problems in using these models: 1) High memory (RAM) usage, 2) High prediction latency, 3) High power dissipation, 4) Poor inference performance on resource constrained devices, and 5) Difficulties in model training [4].

Transformer architecture has become the de-facto deep neural network architecture in sequence-to-sequence modeling tasks in NLP applications. The original implementation consists of an encoder and decoder [5]. In this paper, we focus attention on the encoder part of the transformer that is used in downstream tasks, such as Text Classification and Named Entity Recognition, as implemented in BERT [1]. The architecture encodes information from the text sequence using a sequence of transformer blocks consisting of a self-attention layer followed by a position-wide feed forward layer. Self-attention is one of the most important components in the network and works by refining the representation at each position by aggregating features from all other locations in a single sequence sample. While it has a big advantages in improving model performance, the architecture is over-parametrized.

One of the common methods to simultaneously reduce both the memory footprint and inference time of Deep Neural architectures is to remove some of the model parameters for a marginal drop in performance [5]. These class of methods are referred to as pruning methods. The objective of a pruning technique is to find the order of importance in the model parameters and start removing them, starting from the least important. Pruning also helps alleviate the over parameterization problem in downstream tasks and provides a better generalized model [6]. Pruning methods can be applied after training or while training a machine learning model.

*Work done while A. Parnami was an intern at Wells Fargo during Summer 2021. Corresponding author's email: aparnami@uncc.edu.

The approaches used by pruning algorithms are broadly classified into two categories: a) magnitude based, and b) hessian based [4]. Although hessian-based methods are more accurate, they are not feasible in practice as hessian computations are expensive. Therefore, magnitude-based pruning methods are more popular. As the name implies, parameters are removed solely based on the magnitude of the original weights. Pruning can be performed in a single step or performed iteratively [4].

Although magnitude-based pruning is applied in a lot of architectures, and in applications such as computer vision and NLP, its application to transfer learning is not well studied. Previous papers have investigated improving the model score for applications like machine translation in terms of BLEU score by removing heads in NMT encoder-decoder Transformer. The experiments found that ~90% of heads can be removed without any significant decrease in performance. Another study applied Layerwise relevance propagation (LRP) [6] scores to determine the importance of heads for pruning [7]. LayerDrop is another scheme that attempts to eliminate entire layers from the network [8]. In this paper, we introduce a new technique to prune parameters in a Deep neural network and show empirical results from its application on text classification models based on transformer architecture.

The paper makes the following contributions:

- We apply the *A** search [10] algorithm to prune heads (termed as *A* Pruning*) in downstream models.
- We prune and evaluate the BERT [1] model fine-tuned on text classification tasks.
- We demonstrate on selected datasets that *A** Search prunes ~40% of heads with no drop in model performance and prunes ~%67 of the heads with ~2% drop in accuracy.
- Finally, we demonstrate the efficacy of *A* Pruning* to produce highly compressed downstream models by comparing it with randomized pruning.

The remainder of the paper is organized into the following four sections, background, methodology, experiments and conclusions.

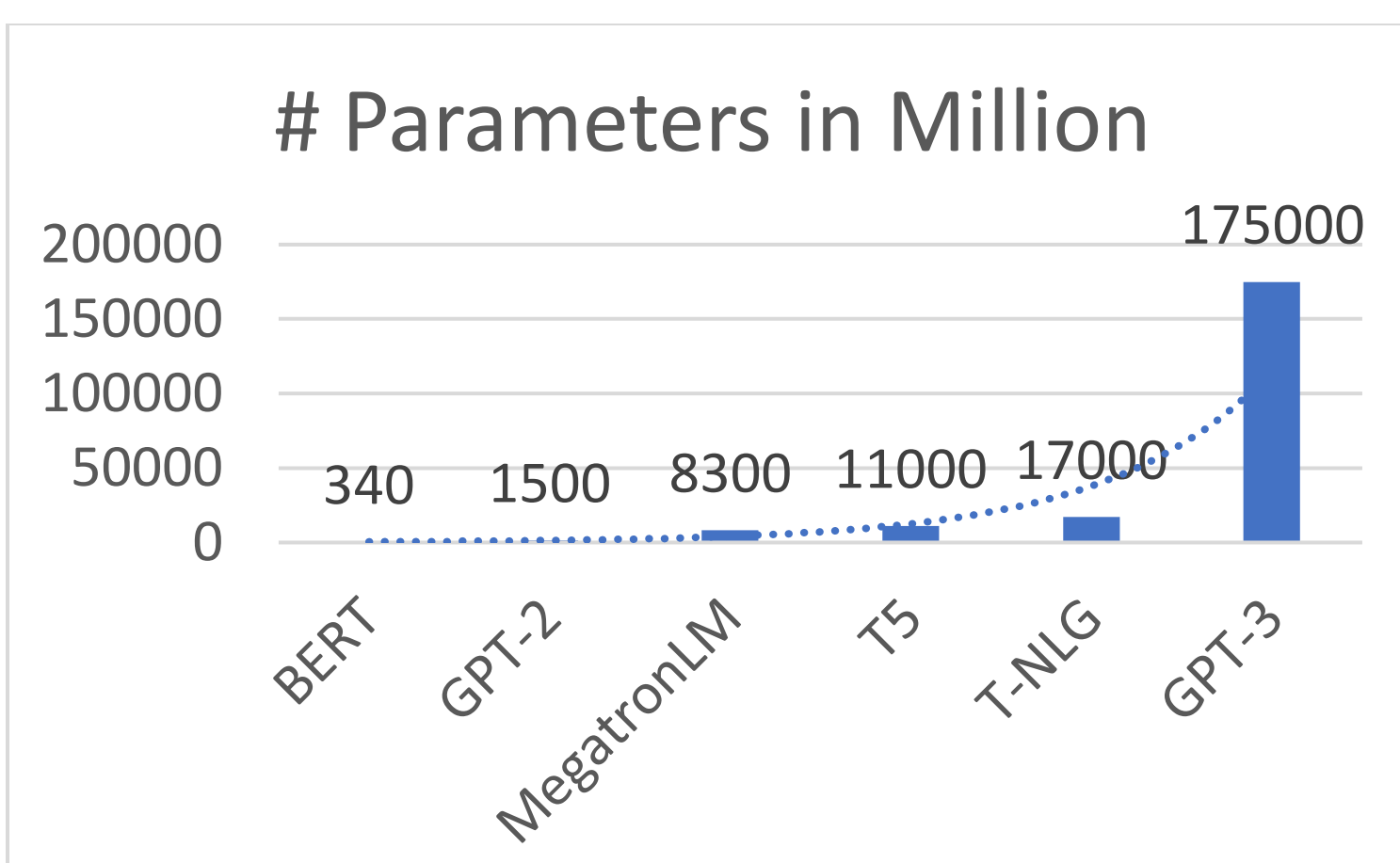

**Figure 1: Growing size of transformer models [4]**

# 2 Background

## 2.1 Model Compression

Compression refers to reducing model size in terms of either number of parameters or amount of storage required to store the model parameters. Gupta & Agarwal [4] classify compression techniques into five methods as shown in Figure 2. This paper focusses on one of the methods called pruning i.e., reducing model size by removing weights (connections) or neurons or layers.

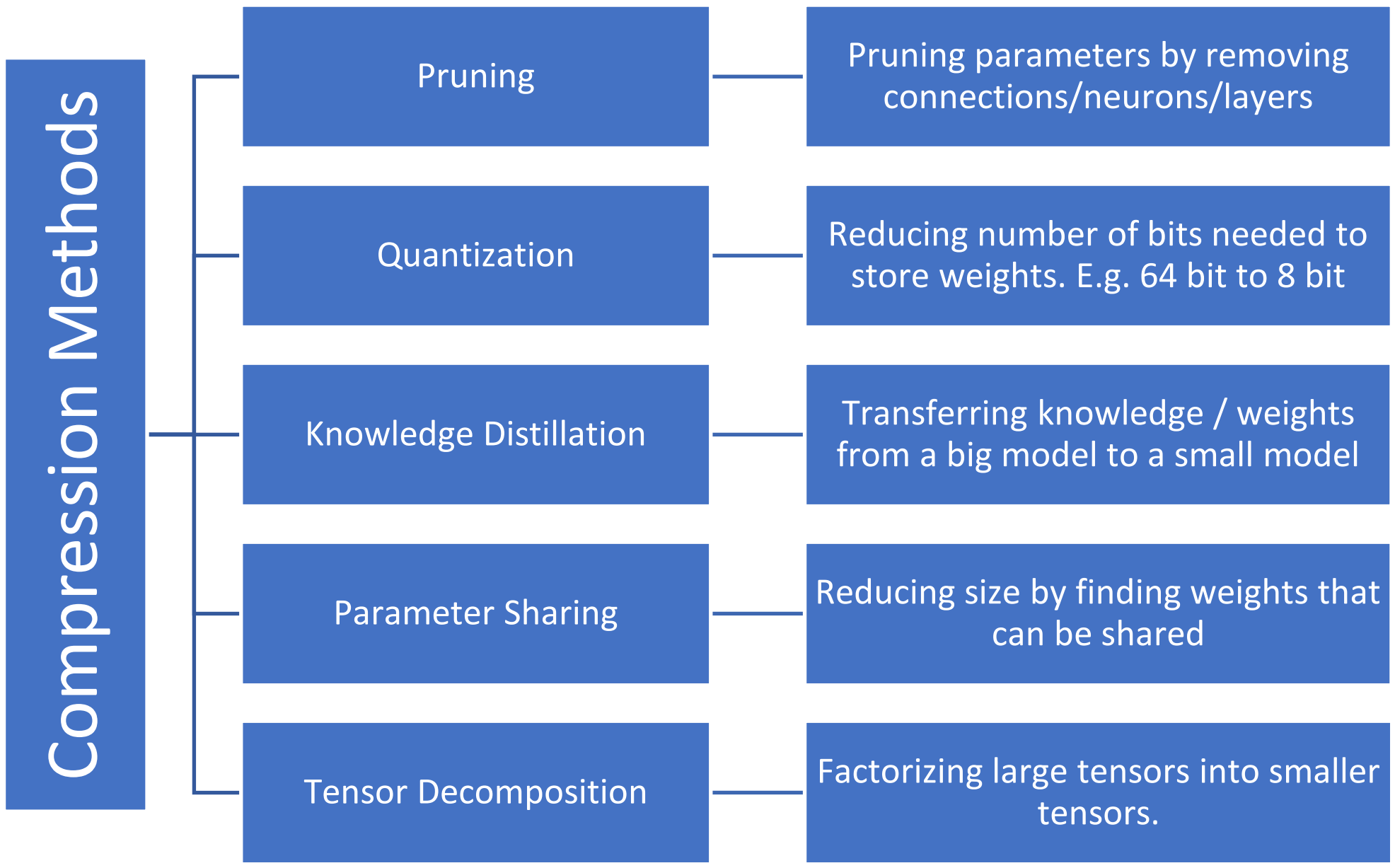

**Figure 2: Compression Methods [4]**

## 2.2 *A** Search

$A*$ (pronounced "A-star") is a graph traversal and path search algorithm, which is often used in many fields of computer science due to its completeness, optimality, and optimal efficiency [9]. $A*$ is an informed search algorithm, or a best-first search, meaning that it is formulated in terms of weighted graphs, and it aims to find least cost path from a source node to a destination node. It does this by maintaining a tree of paths originating at the source node and extending those paths one edge at a time until its termination criterion is satisfied [10].

At each iteration of its main loop, $A*$ needs to determine which of its paths to extend. It does so based on the cost of the path and an estimate of the cost required to extend the path all the way to the goal. Specifically, $A*$ selects the path that minimizes

$$f(n) = g(n) + h(n),$$

where $n$ is the next node on the path, $g(n)$ is the cost of the path from the source node to $n$, and $h(n)$ is a heuristic function that estimates the cost of the cheapest path from $n$ to the goal. $A*$ terminates when the path it chooses to extend is a path from source to destination or if there are no paths eligible to be extended. The choice of the heuristic function is problem specific. If the heuristic function is admissible, meaning that it never overestimates the actual cost to get to the destination, $A*$ is guaranteed to return a least-cost path [10].

# 3 Methodology

## 3.1 Pruning Pipeline

In each of our experiments, we start with a pre-trained BERT [1] model and fine-tune it on our training dataset. The fine-tuned model is then pruned. Next, the pruned model is evaluated on a held-out test set and the steps are shown in Figure 3.

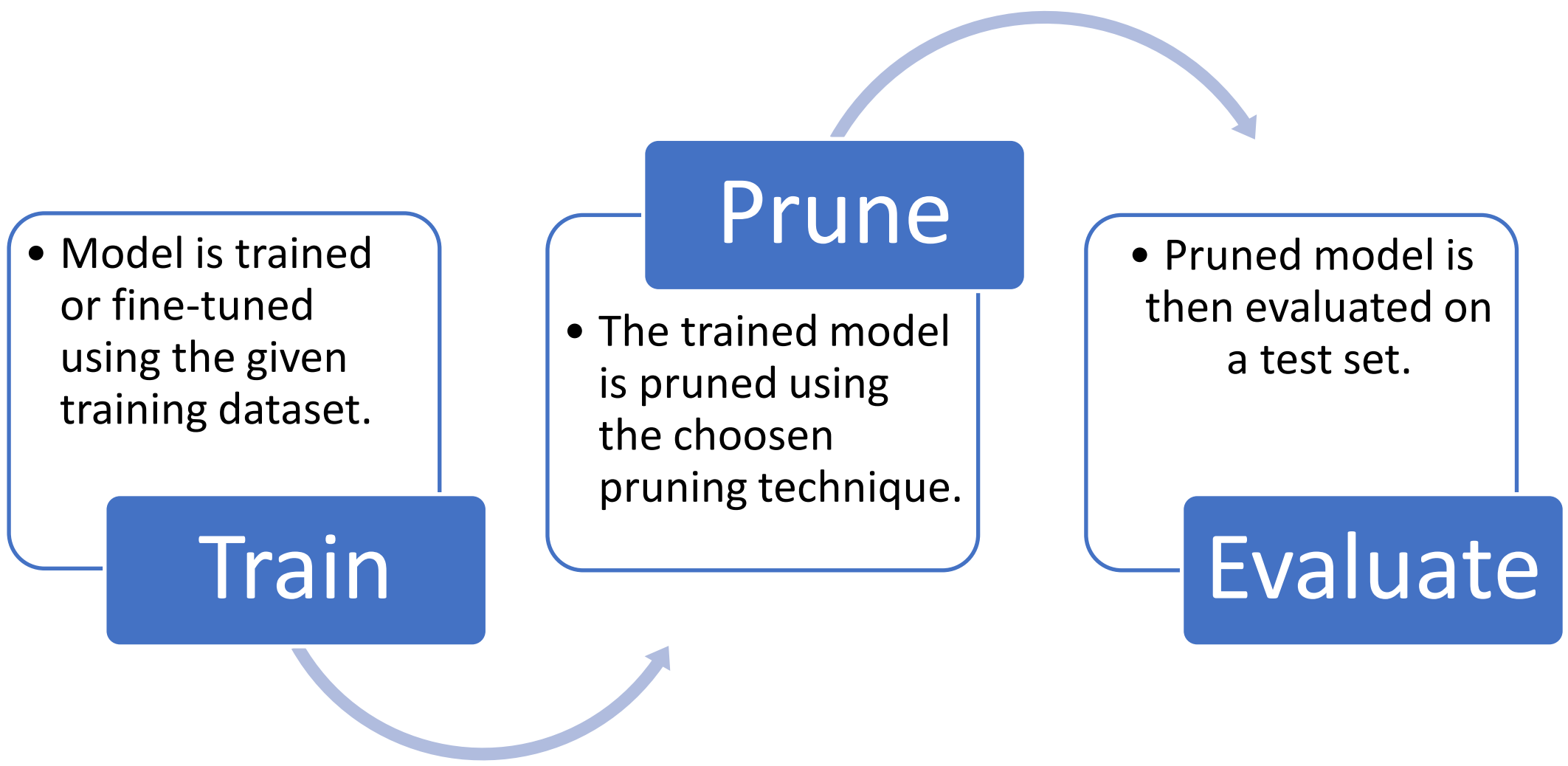

**Figure 3: Pruning Pipeline**

## 3.2 Pruning Self-attention Heads of BERT

BERT (base) [1] has 12 encoders. Each encoder has two types of layers: 1) Dense layers (FFNN) 2) Self-attention layers. Each attention layer has 12 attention heads. This gives us 144 heads in total. Our objective is to prune (remove) heads in attention layers as a way to compress the model without sacrificing too much performance. Two straightforward ways to prune the heads are:

a) **Local Pruning**: A simple way to judge the importance of heads is to remove them one at a time and re-evaluate the model's performance on the testing dataset. Then the head which results in the least drop in accuracy can be pruned away permanently. We refer to this approach as *Local Pruning* (Figure 4a). If there are $l$ attention layers and each layer has $n$ heads, then we have a total of $ln$ heads. Repeating this procedure for all $\boldsymbol{ln}$ heads would require $\frac{\boldsymbol{ln}*(\boldsymbol{ln}+\boldsymbol{1})}{\boldsymbol{2}}$ model re-evaluations to arrive at a desirable solution. Thus, the complexity of Local Pruning is $O(l^2n^2)$ and in case of BERT, it could result in as many as 10,440 searches.

b) **Global Pruning**: Another simple approach is to prune a particular head from every layer and then re-evaluate the model's performance on test dataset. For BERT, this is equivalent to pruning 12 heads at a time. We can then repeat this procedure for all the heads and prune away the head from all the layers which results in the least amount of drop in accuracy. We refer to this approach as *Global Pruning* (Figure 4b). If there are $n$ heads in each attention layer, then the complexity of Global Pruning is $O(n^2)$. Since $l\ == \ n$ in case of BERT-base, $O(n^2) < O(l^2n^2) = O(n^4)$, global pruning is faster than local pruning. On the other hand, removing away a head from all the layers in a single iteration could result in a higher drop in accuracy than pruning away the equivalent number of heads using Local Pruning for the equivalent model compression gains.

This paper presents a novel technique to prune self-attention heads locally but using a guided heuristic to reduce the number of searches. We view pruning as a search problem where the objective is to search for a solution that gives us least drop in accuracy for maximum compression. We use $A*$ search and refer to our approach as *$A*$ Pruning*.

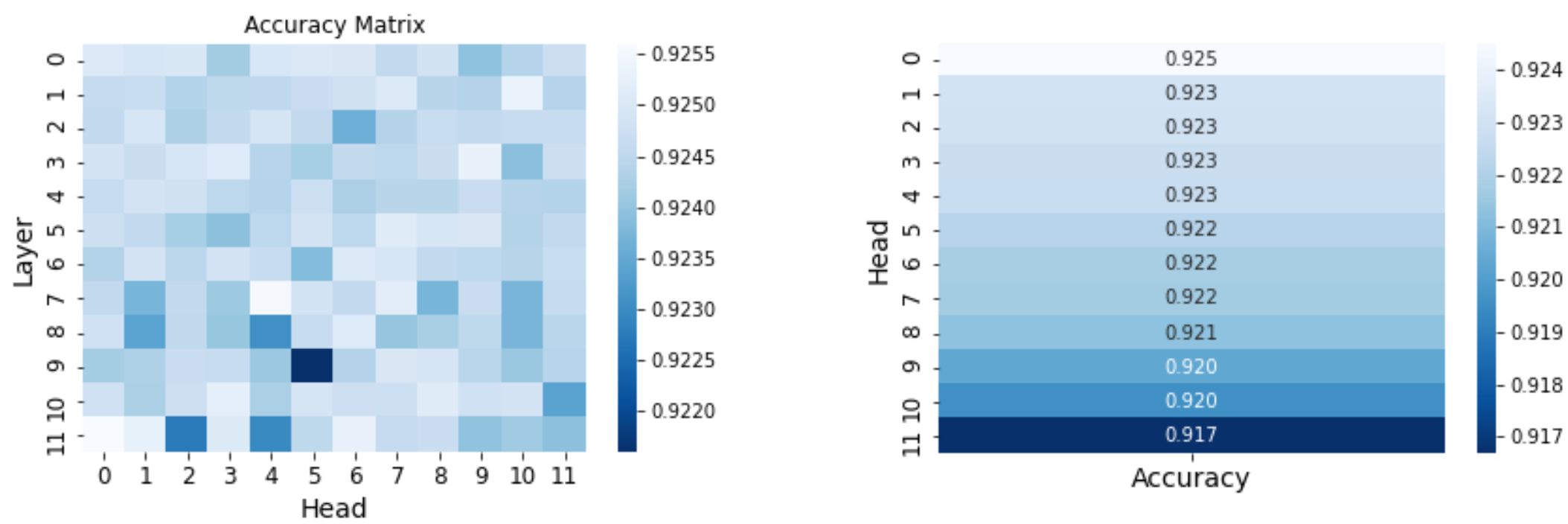

a) Local Pruning

b) Global Pruning

**Figure 4: Local vs Global Pruning: a) Accuracy (post pruning) heatmap when a head at (i,j) is pruned where i is layer and j is head number. b) Accuracy heatmap when a head j is removed from all the layers.**

### 3.2.1 *A* Pruning*

We first discuss the terminology to be used in Section 3.2.2 and then present the algorithm Section 3.2.2.1.

### 3.2.2 *A* Pruning* Variables

To use $A*$ search for pruning, we need to setup some constraint variables that will help us guide the search (Figure 5). These are:

1) **Baseline Accuracy (A):** Model's performance (accuracy) before pruning (Table 2).

2) **Budget (B):** Maximum amount of accuracy we are willing to sacrifice for compression gains.

3) **Accuracy Post Pruning (P):** Model accuracy if one or more of its attention heads were pruned. This is represented as $l \times n$ matrix in case of Local Pruning (Figure 4a).

4) **Cost (C):** The *drop* in performance of the model from A if one or more of its attention heads were to be pruned. This again is represented as $l \times n$ matrix and is computed as $A - P$.

Now, given a budget (B), one can perform local pruning in the following manner:

i. Compute P (Figure 4a)
ii. Compute C
iii. Prune away the head which has the least cost in C
iv. Repeat the procedure (step i - iv) for the remaining unpruned heads.
v. Stop when total cost of pruned heads exceeds the budget B.

However, the above procedure still requires us to recompute the C matrix i.e., cost of pruning remaining heads in each iteration. What if we could only recompute the cost of pruning heads that are most likely going to fit in our budget while not recomputing the higher costing heads? This would reduce the number of reevaluations (searches) in each iteration and help us arrive at an optimal solution faster. To implement this, we define another variable called *Heuristic (H)*.

5) **Heuristic (H):** A method to estimate the cost of pruning a head in the next iteration. For e.g., in iteration 0 we start with P and prune the head with the least cost. In the next iteration 1, we need to know the cost of pruning remaining heads given the first head has been pruned. We use a heuristic (H) to estimate this cost.

**Our choice of Heuristic (H) for *A* Pruning***

- Let the cost of pruning (drop in accuracy) head $(i,j)$ in iteration k = $C^{k}_{(i,j)}$.
- Since in each iteration, a new head is pruned, our assumption is that cost of pruning the same head $(i,j)$ in next iteration is going to increase (i.e., if $(i,j)$ was not pruned in iteration $k$ and another head with least cost was pruned).
- Let the true cost of pruning head $(i,j)$ in iteration k+1 = $C^{k+1}_{(i,j)}$.
- Let the estimated cost (as per the heuristic) of pruning head $(i,j)$ in iteration $k+1$ = $E^{k+1}_{(i,j)}$.

Then, our heuristic (H) estimates the cost of pruning a head in the next iteration to be the ***same as the cost of pruning it in the current iteration*** i.e.,

$$E^{k+1}_{(i,j)} = C^{k}_{(i,j)}$$

Under the assumption that pruning results in loss in accuracy or increase in cost, we can say that $C^{k}_{(i,j)} < C^{k+1}_{(i,j)}$ and hence the estimated cost is always going to be less than the true cost:

$$E^{k+1}_{(i,j)} < C^{k+1}_{(i,j)}$$

This assures us that our heuristic will not overshoot the true cost and hence will not eliminate excessive number of heads during search.

6) **Search Space (S):** This is the number of heads that we look at in each iteration i.e., those that have not been pruned or eliminated by $A*$ search. In the beginning, this is equal to $l * n$.

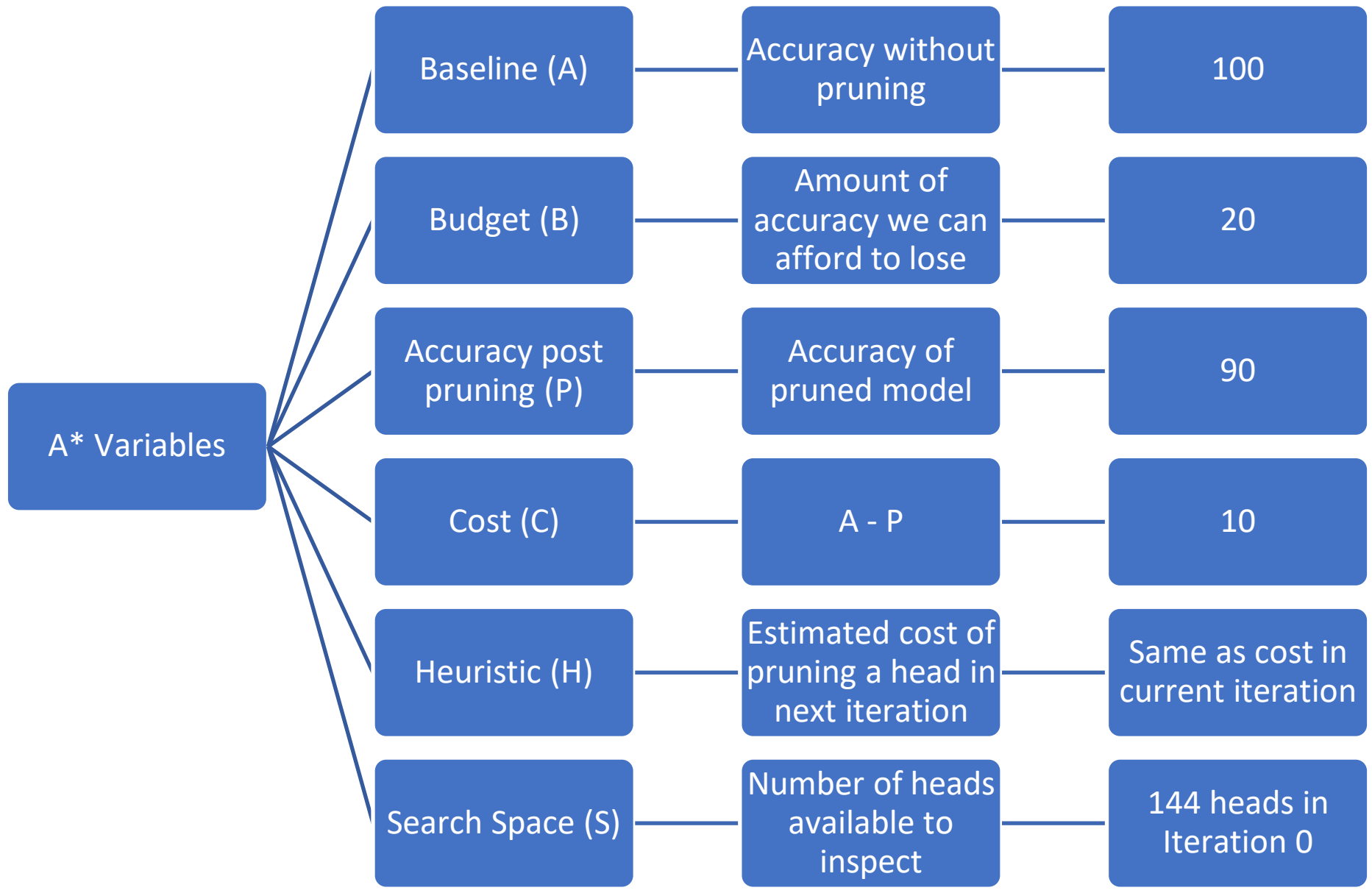

**Figure 5: *A** Variables**

### 3.2.2.1 The Algorithm

The input to our *A* Pruning* algorithm (Figure 6) are the *A** variables defined above. The baseline accuracy of the unpruned model is given as A, the budget is set to B, and the matrix P and C are just initialized but not computed yet. S is initialized to $l * n$ and the heuristic (H) is given. Let L represent the output solution i.e., list of heads to be pruned.

Figure 6 shows the details of the algorithm. It works in an iterative process and removes the lowest performing, head in a single iteration. At each iteration, we recalculate the heuristics to find the search space for possible heads that can be removed without exceeding the Budget (B). We stop the pruning algorithm when we have crossed the budget. At the end of pruning process, we have a list of heads that can be pruned without crossing the budget.

Figure 7 provides an example that demonstrates *A* Pruning*.

**ALGORITHM FOR PRUNING HEADS USING *A** SEARCH**

**Input**

| | |
|---|---|
| **A** | *Accuracy of the model when none of its head is pruned.* |
| **B** | *Budget i.e., The amount of accuracy we can afford to lose.* |
| **P** | *Post pruning accuracy of $l \times n$ heads can be represented by a matrix P.* |
| **C** | *C is a $l \times n$ matrix representing the drop in accuracy from the baseline.* |
| **S** | *Number of heads that have not been pruned or eliminated by A* Search.* |
| **H** | *Estimated cost of pruning a head in next iteration.* |

**Output**

| | |
|---|---|
| **L** | *List of heads pruned i.e. the solution of A* Pruning.* |

**BEGIN**

```
SET B
SET S = l * n
WHILE B > 0 and S > 0
    Calculate P                  // Post pruning accuracy of unpruned heads
    Calculate C                  // Compute cost of the unpruned heads
    C = SORT (C)                 // Sort cost C of unpruned heads in the ascending order
    X = ARGMIN (C)               // head with the lowest cost
    C_x = MIN (C)                // Cost of X
    IF C_x < 0 THEN
        SET C_x = 0
    IF C_x < B THEN
        SET X as pruned
        APPEND X to L
        UPDATE B = B – C_x
        UPDATE S = S – 1
        IF S > 0 THEN
            SET T = 0            // Total estimated cost of pruning remaining heads in S
            FOR EACH HEAD y in S // Iterating over remaining heads sorted by increasing cost
                Calculate E_y using H   // The estimated (H) non-negative cost of pruning y given heads in L have been pruned
                I_y = E_y – C_x         // Estimated contribution of head y to cost E_y
                IF T + I_y < = B THEN
                    T = T + I_y
                ELSE
                    SET y as eliminated
                    SET remaining unvisited heads as eliminated
                    LET K = # remaining unvisited heads + 1
                    Update S = S – K
                    BREAK
    ELSE
        BREAK
```

**END**

**Figure 6: *A* Pruning* algorithm**

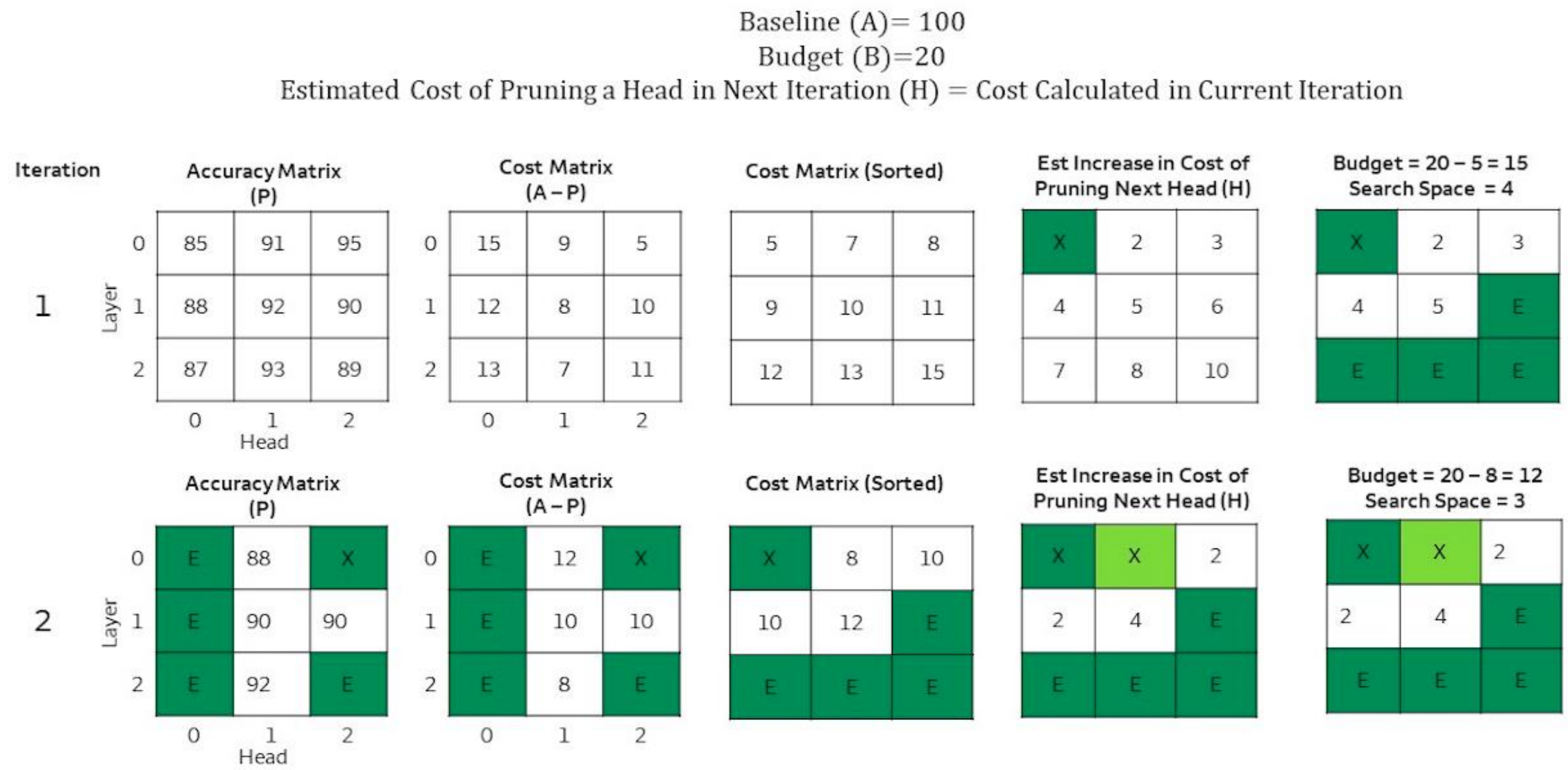

**Figure 7:** ***A* Pruning*** **Example:** X represents a pruned head and E represents a head eliminated by *A** search. In iteration 1, the least cost of pruning a head is 5, so the head corresponding to this cost (X) is pruned. The available budget is 20 – 5 = 15. With this budget we could fit the heads with estimated costs of 2,3,4,5 as their sum 14 is within the budget. The remaining high costing heads are eliminated from search.

## 4 Experiments

### 4.1 Datasets

We illustrate the results on two datasets for text classification using a BERT-base model. The text classification uses the feature representation of [CLS] token [1] that is further passed to a linear layer to get a prediction score for the classes. We train and evaluate the model's pruning performance on Amazon [11] and IMDB [12] reviews datasets. Both dataset have binary labels i.e., text reviews labelled as positive or negative. Amazon dataset has 200K training samples and 25K testing samples whereas IMDB dataset has 25K training samples and 25K testing samples. Both the datasets are balanced. (Table 1)

| **Dataset** | **Classes** | **Type** | **Average Lengths** | **Max lengths** | **Train Samples** | **Test Samples** |
|---|---|---|---|---|---|---|
| Amazon | 2 | Sentiment | 244 | 5019 | 200,000 | 25,000 |
| IMDB | 2 | Sentiment | 294 | 3541 | 25,000 | 25,000 |

**Table 1: Datasets**

### 4.2 Target Model

We use BERT-base [13] as our Transformer model of choice in these experiments. BERT has 12 encoders, and each of the encoders has two types of layers: 1) Dense layers and 2) Self-attention layers. We perform pruning of self-attention heads and analyze the effect.

### 4.3 Baseline Accuracy: Before Compression

Before pruning, we evaluate the performance of the BERT-base on the task of sentiment analysis using both Amazon and IMDB datasets. We start with a pre-trained model and fine-tune it on each of the datasets. The implementation of BERT-base is obtained from Hugging Face [14] library and Adam [15] optimizer is used in the process of fine-tuning. The baseline accuracy and details of hyperparameters are given in Table 2.

| **Dataset** | **Test Accuracy (%)** | **Hyperparameters** | | | |
|---|---|---|---|---|---|
| | | **Embedding Seq Length** | **Batch Size** | **Learning Rate** | **Finetune Epochs** |
| Amazon | 92.46 | 64 | 64 | $10^{-5}$ | 2 |
| IMDB | 88.14 | 128 | 64 | $10^{-5}$ | 2 |

**Table 2: Baseline Accuracy**

### 4.4 Results

#### 4.4.1 ***A* Pruning***

We report the results for *A* Pruning* in Table 3. For each of *A* Pruning* experiments, we allocate a maximum budget **B**. We report the budget used as **X.** Interestingly, for all the experiments in Table 3, the *A* Pruning* method could find heads that contributed negatively to model's performance i.e., when pruned, model's accuracy actually improved (Figure 8c). Therefore, our budget utilization is zero for such heads. In fact, we found that we could prune **~40%** of heads without any loss in accuracy. This indicates that our baseline model is potentially over-parametrized for the task of sentiment analysis and there is a good scope for compression.

For Amazon dataset with the given budget of 1, *A* Pruning* can eliminate 76 heads (out of 144) i.e., achieve ~53% compression with only 0.5 loss in accuracy. With a budget of 2, the % compression is ~60% and with budget of 3, it is ~67%. Obviously, as we increase the budget, the % compression increases. However, increasing the budget also increases the available search space (the set of solutions among which the desired solution resides) during the *A** search. We observed similar effects on IMDB dataset (Table 3).

| **Dataset** | **Budget** | | | **Heads Pruned (#), Compression (%), Accuracy Post Pruning (P)** | | | | | | **# Search** |
|---|---|---|---|---|---|---|---|---|---|---|
| | Given (B) | Used (X) | Remaining (B-X) | With zero budget | | | With budget (X) | | | |
| | | | | # | % | P | # | % | P | |
| Amazon | 1 | 0.5 | 0.5 | 59 | 40.97 | 92.46 | 76 | 52.77 | 91.96 | 5282 |
| | 2 | 1.59 | 0.41 | 57 | 39.58 | 92.46 | 86 | 59.72 | 90.87 | 6360 |
| | 3 | 2.72 | 0.28 | 60 | 41.66 | 92.46 | 96 | 66.66 | 89.74 | 7496 |
| IMDB | 1 | 0.04 | 0.96 | 58 | 40 | 88.14 | 66 | 51.76 | 87.28 | 6158 |
| | 2 | 1.42 | 0.58 | 58 | 40.27 | 88.14 | 76 | 52.77 | 86.72 | 6420 |
| | 3 | 2.6 | 0.4 | 60 | 41.66 | 88.14 | 86 | 59.72 | 85.54 | 7107 |

**Table 3: Results for *A* Pruning***

We also present results visually for Amazon dataset with budget = 1 in Figure 8 and the remaining visual results are presented in Appendix (*A* Pruning* Results).

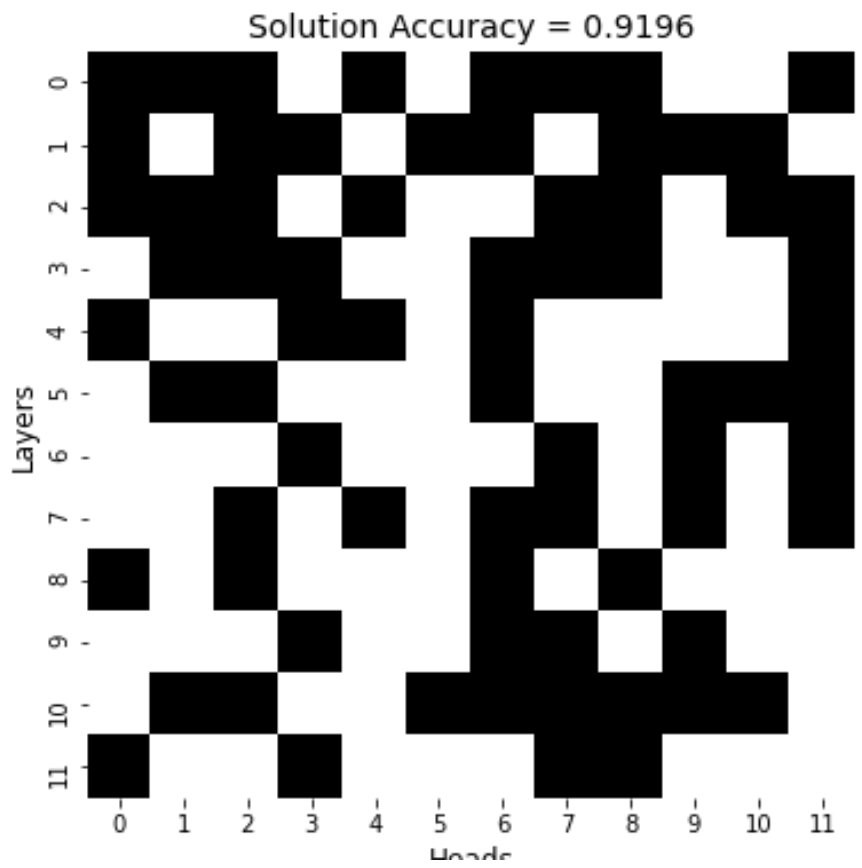

**a) Final Solution**

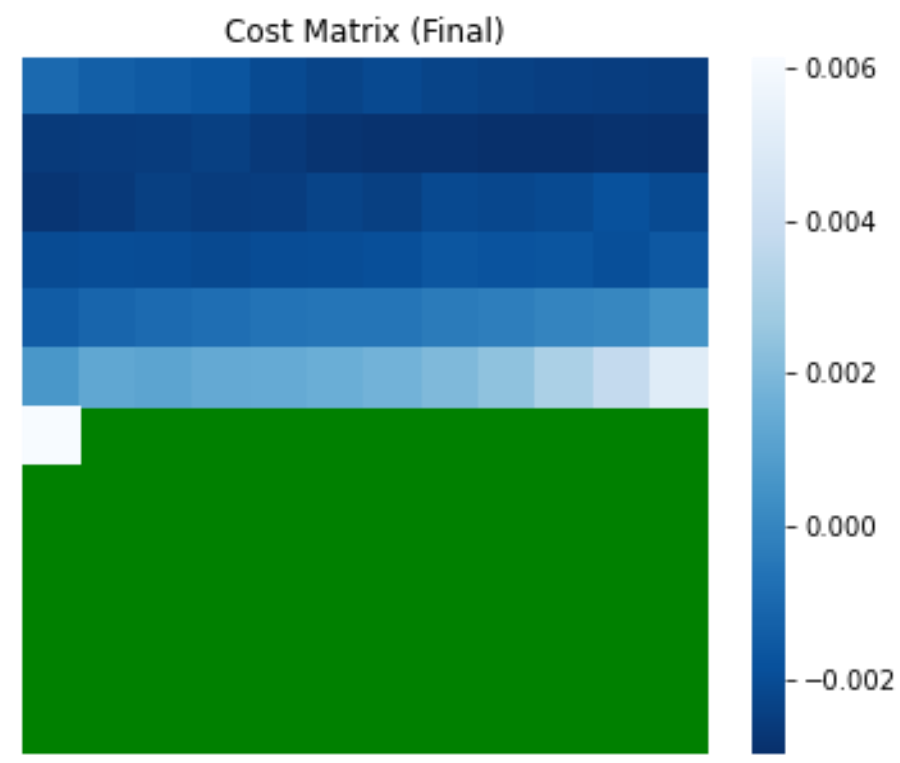

**b) Cost Matrix**

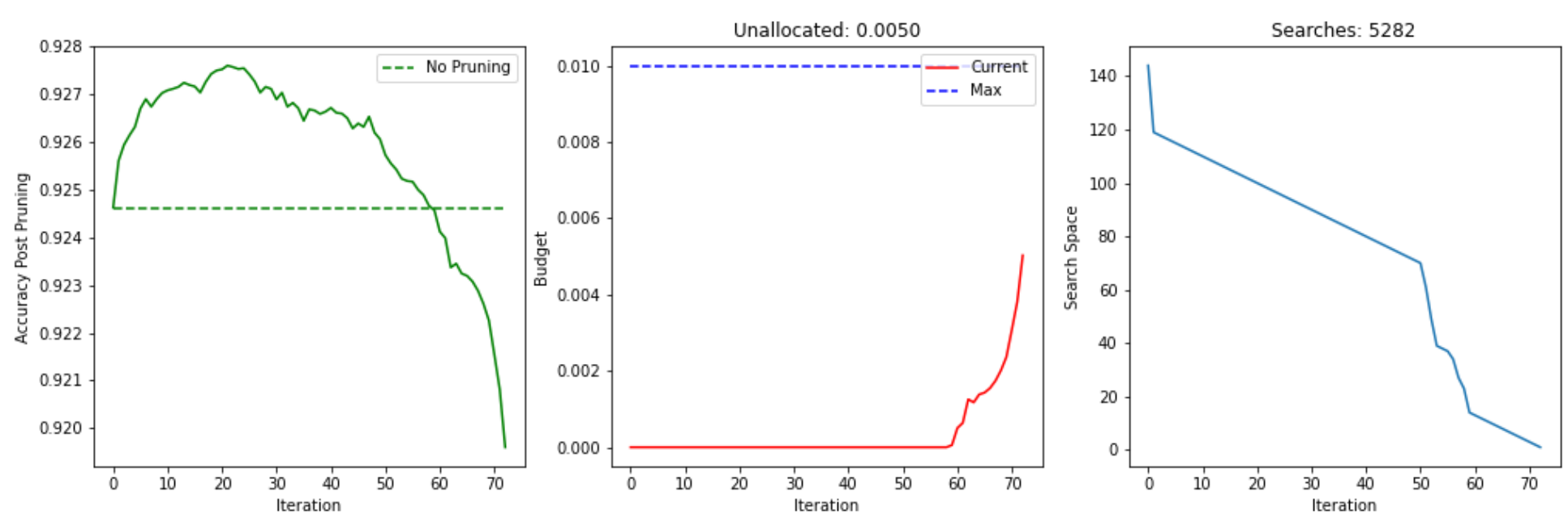

**c) Accuracy (left), Budget used (center) and # Searches (right) vs Iteration**

**Figure 8: *A* Pruning* results for Amazon dataset with a budget = 1. a)** Final solution presented as layer (y) vs attention head (x) matrix. Heads are represented as pruned (black) and unpruned (white). **b)** The sorted cost matrix (C) shows the cost of pruned heads (in blue) and it also shows the heads that were eliminated by the heuristic used in *A* Pruning* (green). **c)** Accuracy vs Iteration (left) represents the change in accuracy as the heads are pruned away in each iteration. Similarly, budget used (center) and the number of heads searched (right) in each iteration.

Figure 8c, captures the behavior of *A* Pruning*. At the beginning (before pruning) the model's baseline accuracy is represented as a dotted line in Figure 8c (left). At this point the budget used is zero and our search space includes all heads (144). In each iteration, the algorithm finds a least cost head and prunes it. It also updates its search space based on the available budget and the heuristic (H). The accuracy increases as we begin pruning and reaches a peak. It then descends and drops back to the baseline accuracy. During this period the utilized budget is zero. We also observe that the cost of pruning most heads during this period is negative, which explains the linear trend in number of searches i.e., the *A** heuristics does not kick in until the cost becomes positive. Once it does, the *A** algorithm starts eliminating heads, which explains the drop in number of searches (Figure 8c (right).)

### 4.4.2 Random Pruning

We establish a baseline for compression by comparing the result of *A* Pruning* to Random Pruning i.e., if given the same budget (B), how many heads at maximum can be pruned if the pruning strategy is to pick the heads randomly. For a given budget, we perform 100 pruning trials. Figure 9 shows the results for the Amazon dataset with a budget (B) of 3. The top figure (green) presents the distribution of the number of heads pruned in the 100 trials and the bottom figure presents the distribution of the budget used. Note that the solution found by *A* Pruning* dramatically outperforms Random Pruning.

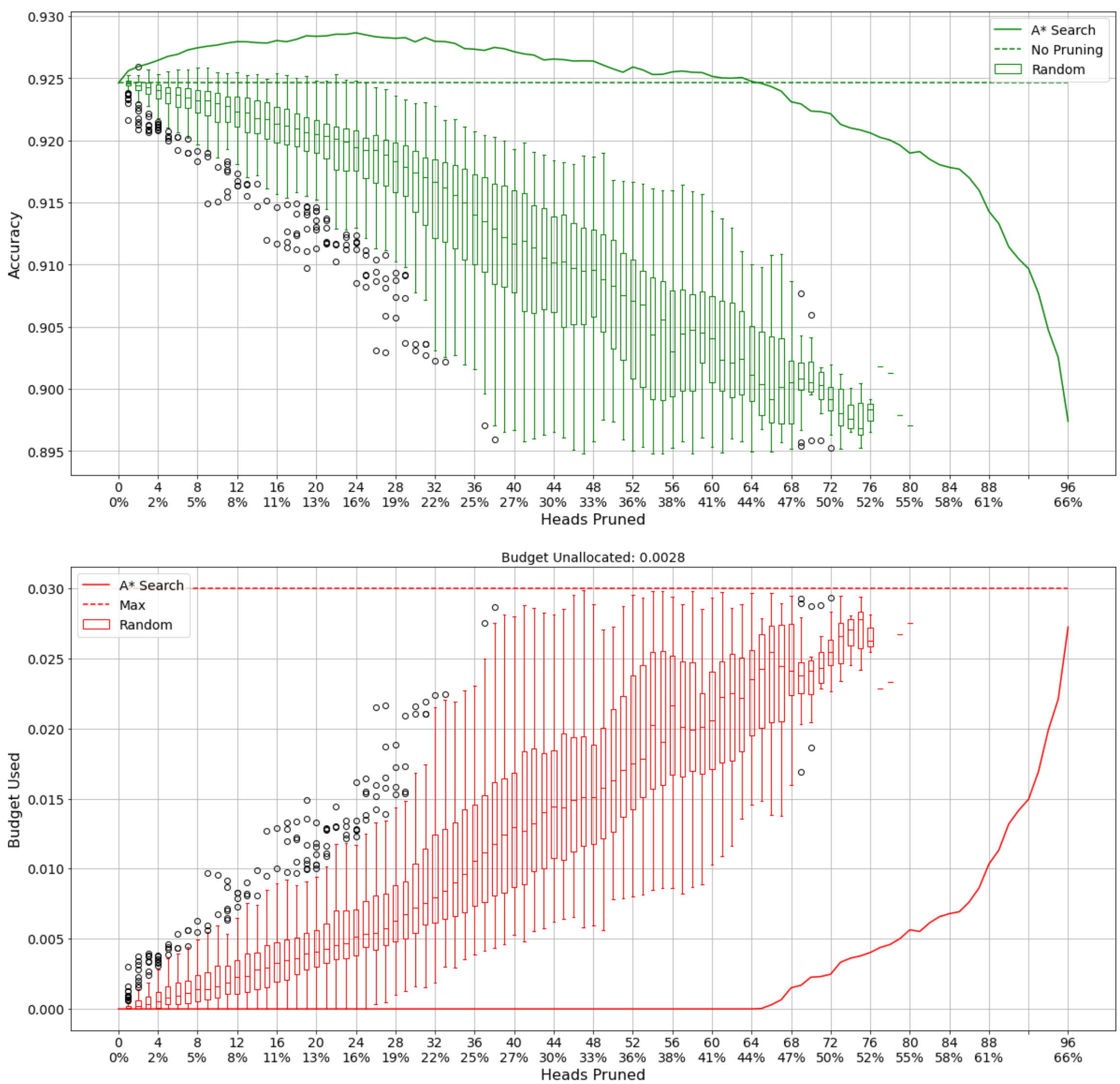

**Figure 9: Random vs *A** Pruning on Amazon dataset with budget 3.**

**Top: Accuracy vs # Heads Pruned (Compression %).**

**Bottom: Budget Used vs # Heads Pruned (Compression %)**

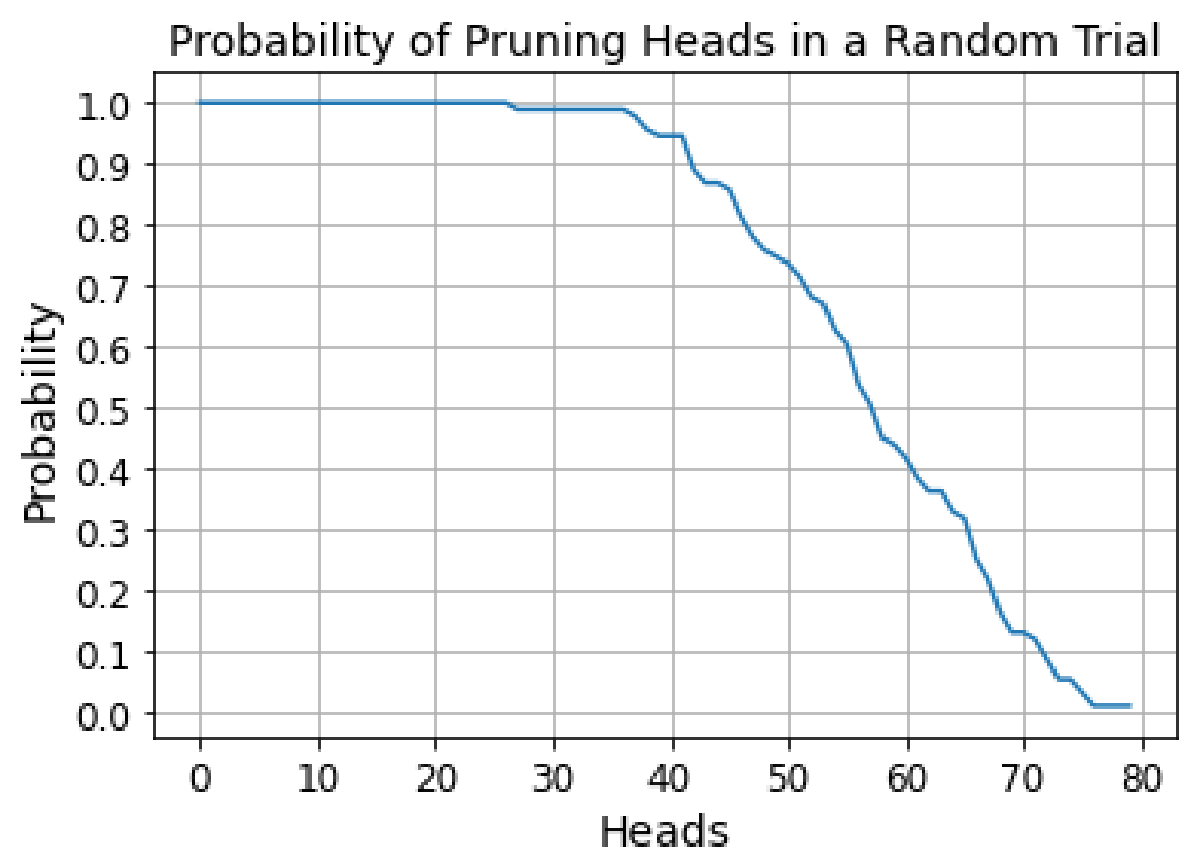

**Figure 10: Probability of pruning x heads in a random trial on amazon dataset with budget 3**

A random pruning trial is most likely to result in pruning of approximately 30 heads (Figure 10) in comparison to 96 heads eliminated by *A* Pruning* (Table 3). Moreover, the probability of pruning more than 30 heads decreases thereafter, and it is almost zero for pruning more than 80 heads (Figure 10). This demonstrates that we cannot just prune any head.

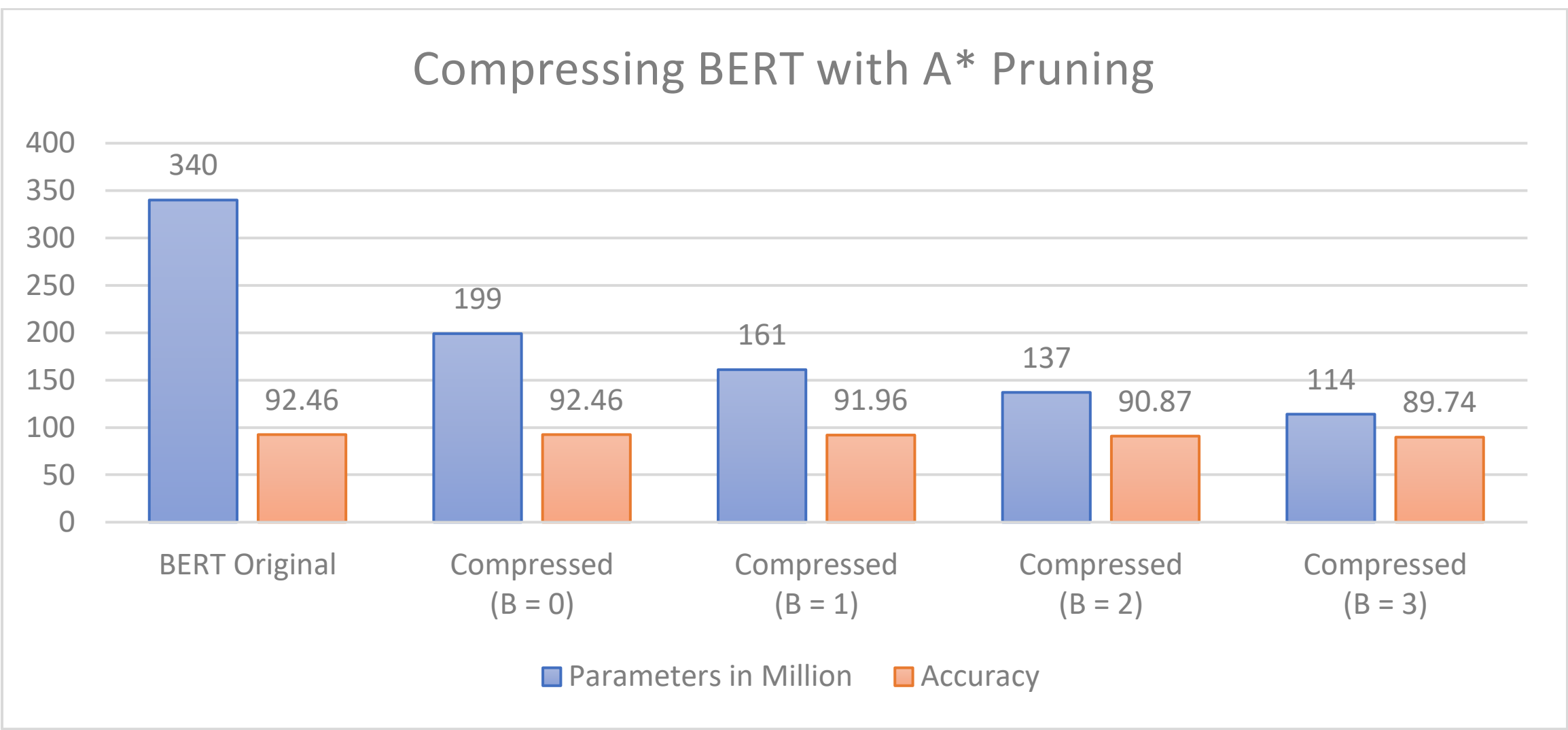

**Figure 11: Parameters in Million vs Accuracy of BERT model compressed using *A* Pruning* with budget B.**

Figure 11 compares the compressed models found by *A* Pruning* to the original BERT model. We can observe a significant reduction in the number of parameters for little to no drop in accuracy.

## 5 Conclusion

This paper presented a novel approach for pruning attention heads of Transformer models such as BERT. The *A* Pruning* approach uses *A** search to prune least significant heads in the model, where the significance of a head can be defined by a performance metric such as accuracy. Moreover, by combining *A** search with a budget, our approach guarantees to deliver a compressed model whose accuracy will not drop beyond the set budget. We evaluated our approach on the task of sentiment analysis using Amazon and IMDB datasets and found that *A* Pruning* could remove ~40% of the heads from the BERT model with no loss in accuracy.

## References

[1] J. Devlin, M. K. Chang, K. Lee and K. Toutanova, "Bert: Pre-training of deep bidirectional transformers for language understanding," *arXiv preprint,* vol. arXiv:1810.04805, 2018.

[2] A. Radford, J. Wu, R. Child, D. Luan, D. Amodei and I. Sutskever, "Language Models are Unsupervised Multitask Learners," *arxiv,* 2018.

[3] T. Brown, B. Mann, N. Ryder, M. Subbiah, J. D. Kaplan, P. Dhariwal, A. Neelakantan, P. Shyam, G. Sastry, A. Askell, S. Agarwal, A. Herbert-Voss and G. Krueger, "Language Models are Few-Shot Learners," *NEURIPS,* vol. 33, pp. 1877-1901, 2020.

[4] M. Gupta and P. Agrawal, "Compression of Deep Learning Models for Text: A Survey," *TKDD,* 2020.

[5] A. Vaswani, N. Shazeer, N. Parmar, J. Uszkoreit, L. Jones, A. N. Gomez, L. Kaiser and I. Polosukhin, "Attention Is All You Need," *NIPS 2017,* 2017.

[6] M.-T. L. a. C. D. M. Abigail See, "Compression of neural machine translation models via pruning," *arXiv:1606.09274,* 2016.

[7] L. Arras, F. Horn, G. Montavon, K. Muller and W. Samek, ""What is Relevant in a Text Document?": An Interpretable Machine Learning Approach," *PLOS ONE,* vol. 12, 2017.

[8] E. Voita, D. Talbot, F. Moiseev and R. a. T. I. Sennrich, "Analyzing multi-head self-attention: Specialized heads do the heavy," *arXiv:1905.09418,* 2019.

[9] A. Fan and E. a. J. A. Grave, "Reducing transformer depth on demand with structured dropout," *arXiv:1909.11556,* 2019.

[10] P. E. Hart, N. J. Nilsson and B. Raphael, "A Formal Basis for the Heuristic Determination of Minimum Cost Paths," *IEEE Transactions on Systems Science and Cybernetics,* vol. 4, pp. 100-107, 1968.

[11] S. J. Russell and P. Norvig, Artificial Intelligence: A Modern Approach, Boston: Pearson, 2018.

[12] "A* Search Algorithm," [Online]. Available: https://en.wikipedia.org/wiki/A*_search_algorithm#Description.

[13] J. a. J. L. McAuley, "Hidden factors and hidden topics: understanding rating dimensions with review text.," *Proceedings of the 7th ACM conference on Recommender systems,* 2013.

[14] A. L. a. D. R. E. a. P. P. T. a. H. D. a. N. A. Y. a. P. C. Maas, "Learning Word Vectors for Sentiment Analysis," in *Proceedings of the 49th Annual Meeting of the Association for Computational Linguistics: Human Language Technologies*, Portland, Oregon: Association for Computational Linguistics, 2011, pp. 142--150.

[15] J. M.-W. C. K. L. a. K. T. Devlin, "Bert: Pre-training of deep bidirectional transformers for language understanding," *arXiv preprint,* 2018.

[16] T. Wolf, L. Debut, V. Sanh, J. Chaumond, C. Delangue, A. Moi, P. Cistac, T. Rault, R. Louf, M. Funtowicz and J. Brew, "HuggingFace's Transformers: State-of-the-art Natural Language Processing," *ArXiv,* vol. abs/1910.03771, 2019.

[17] D. P. Kingma and J. Ba, "Adam: A method for stochastic optimization.," *arXiv preprint arXiv:1412.6980,* 2014.

[18] R. Rudinger, J. Naradowsky, B. Leonard and B. V. Durme, "Gender Bias in Coreference Resolution," *Proceedings of the 2018 Conference of the North American Chapter of the Association for Computational Linguistics: Human Language Technologies, Volume 2 (Short Papers),* pp. 8-14, 2018.

[19] M. Zhu and S. Gupta, "To prune, or not to prune: exploring the efficacy of pruning for model compression," *arXiv:1710.01878,* 2017.

[20] Z. Wang and J. a. L. T. Wohlwend, "Structured Pruning of Large Language Models," *arXiv:1910.04732,* 2019.

[21] V. Sanh, T. Wolf and A. M. Rush, "Movement Pruning: Adaptive Sparsity by Fine-Tuning," *arXiv:2005.07683,* 2020.

[22] A. Paszke, S. Gross, S. Chintala, C. Gregory, E. Yang, Z. DeVito, L. Zeming, A. Desmaison, L. Antiga and A. Lerer, "Automatic differentiation in Pytorch," *NIPS-W,* 2017.

[23] P. Michel and O. a. N. G. Levy, "Are Sixteen Heads Really Better than One?," *arXiv:1905.10650,* 2019.

[24] T. Wolf, L. Debut, V. Sanh, J. Chaumond, C. Delangue, A. Moi, P. Cistac, T. Rault, R. Louf, M. Funtowicz, J. Davison, S. Shleifer, P. v. Platen, C. Ma, Y. Jernite, J. Plu, C. Xu and L. ScaoTeven, "HuggingFace's Transformers: State-of-the-art Natural Language Processing," *arXiv:1910.03771,* 2019.

# Appendix

## 1 *A* Pruning* Results

In this section, we present the visual results for the experiments described in Table 3.

### 1.1 Amazon with Budget = 2

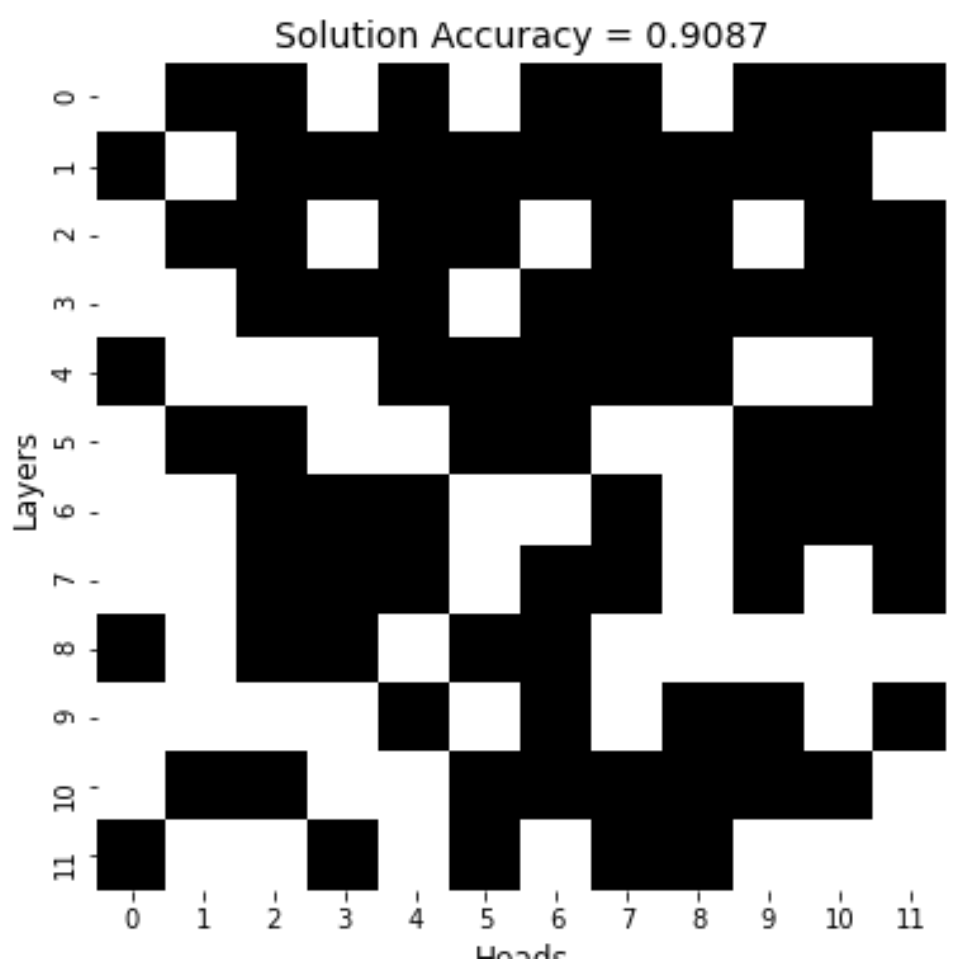

**d) Final Solution**

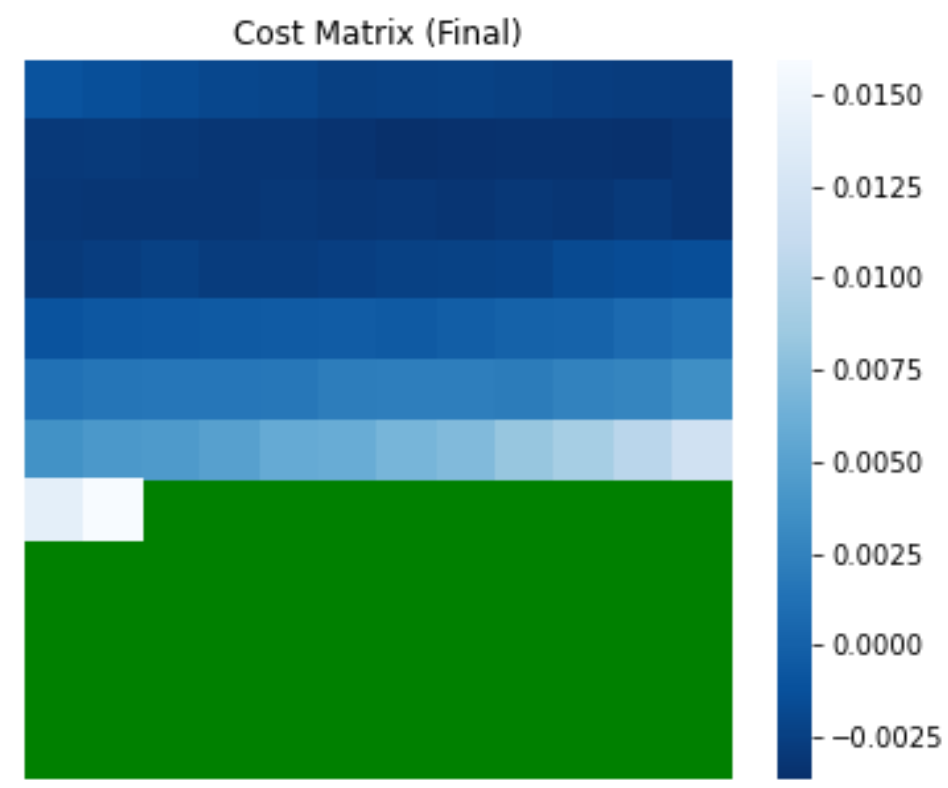

**e) Cost Matrix**

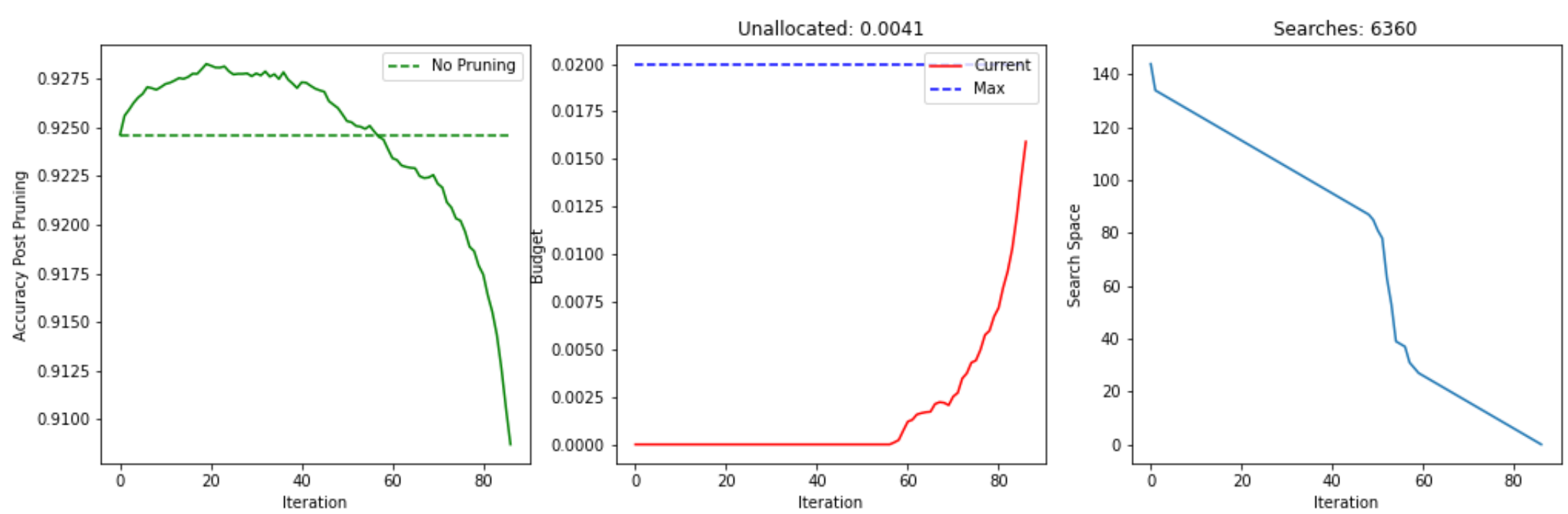

**f) Accuracy (left), Budget used (center) and # Searches (right) vs Iteration**

**Figure 12: *A* Pruning* results for Amazon dataset with a budget = 2. a) Final solution presented as layer (y) vs attention head (x) matrix. Heads are represented as pruned (black) and unpruned (white). b) The sorted cost matrix (C) shows the cost of pruned heads (in blue) and it also shows the heads that were eliminated by the heuristic used in *A* Pruning* (green). c) Accuracy vs Iteration (left) represents the change in accuracy as the heads are pruned away in each iteration. Similarly, budget used (center) and the number of heads searched (right) in each iteration.**

### 1.2 Amazon with Budget = 3

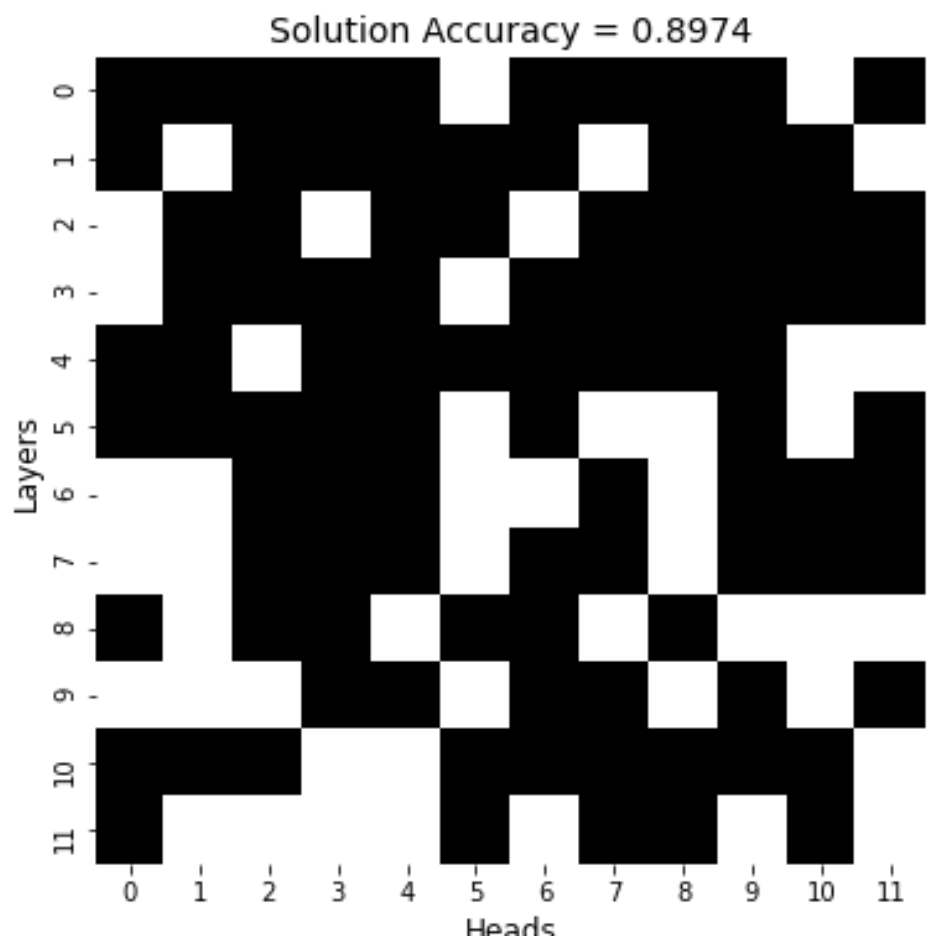

**a) Final Solution**

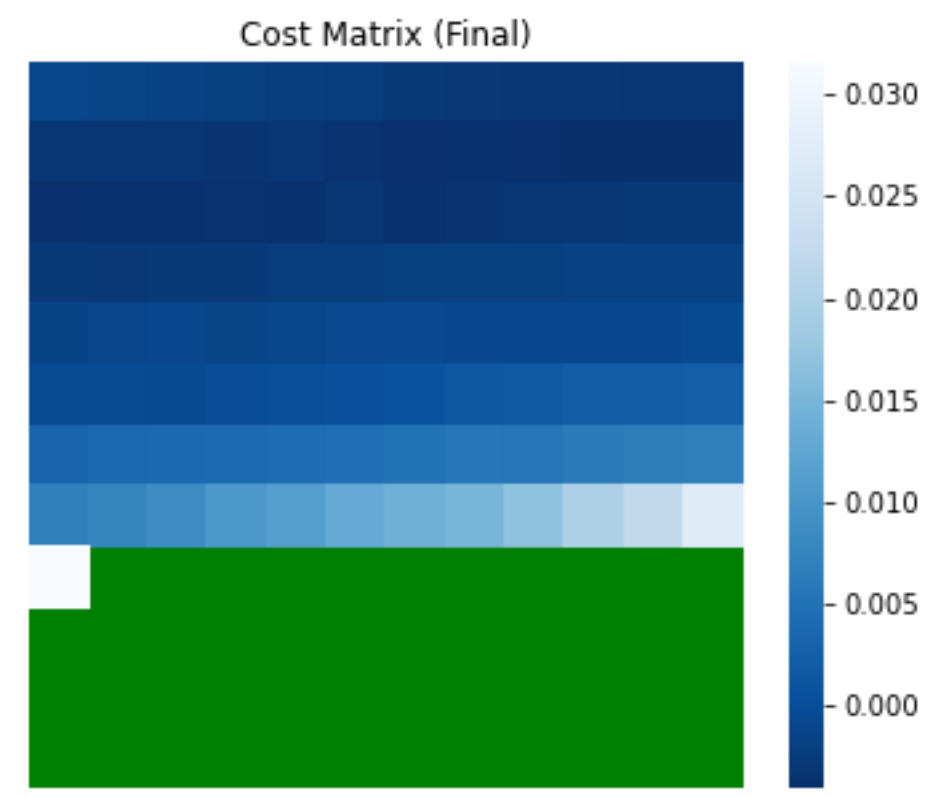

**b) Cost Matrix**

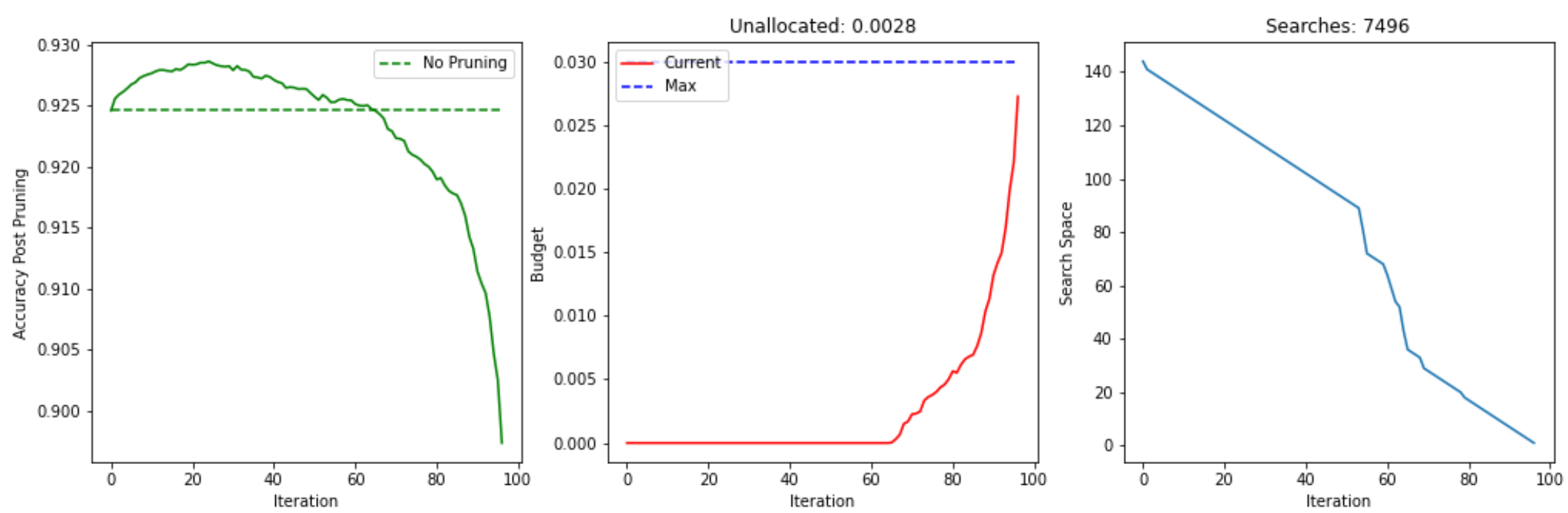

**c) Accuracy (left), Budget used (center) and # Searches (right) vs Iteration**

**Figure 13: Results on Amazon dataset with budget = 3**

## 1.3 IMDB with Budget = 1

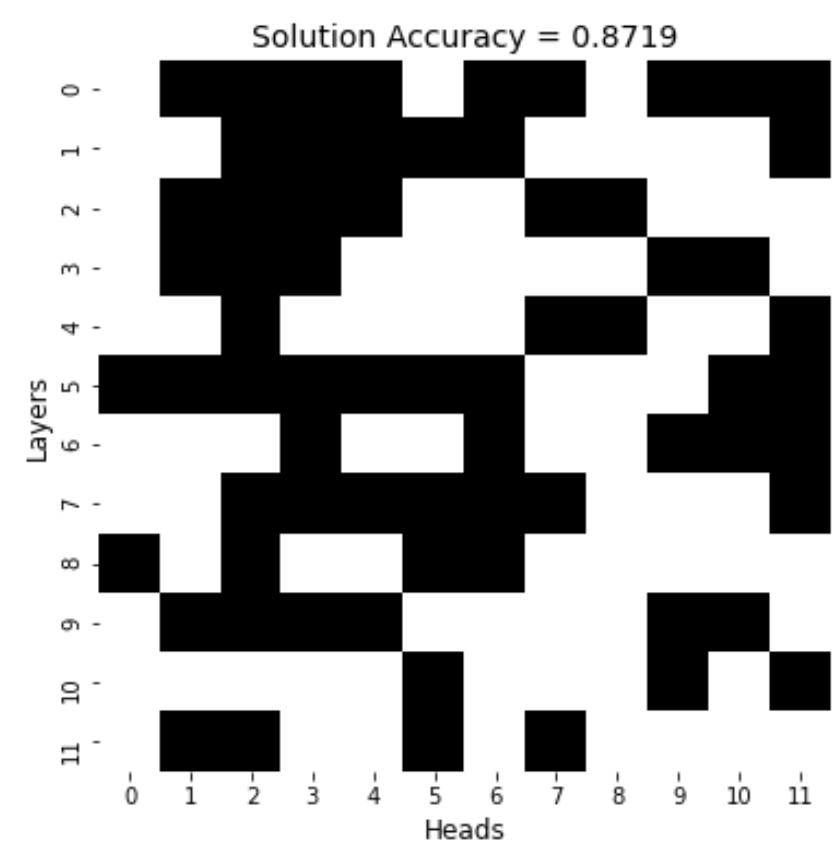

**a) Final Solution**

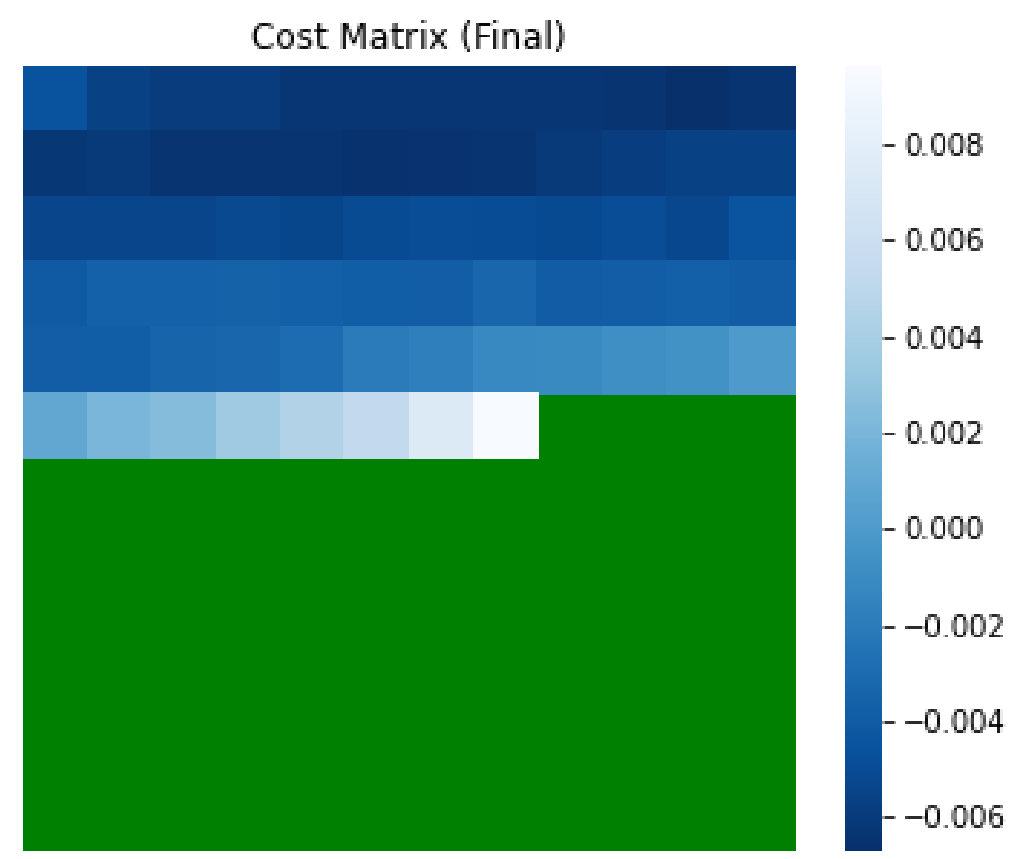

**b) Cost Matrix**

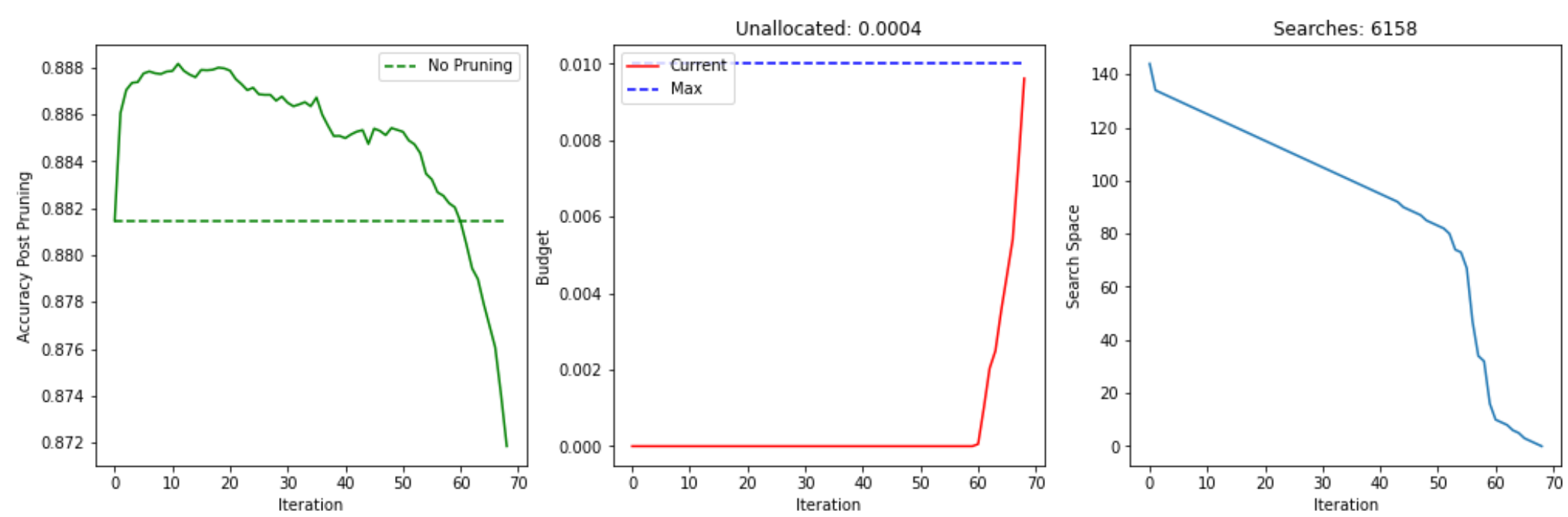

**c) Accuracy (left), Budget used (center) and # Searches (right) vs Iteration**

**Figure 14: *A** Pruning results on IMDB dataset with Budget = 1**

### 1.4 IMDB with Budget = 2

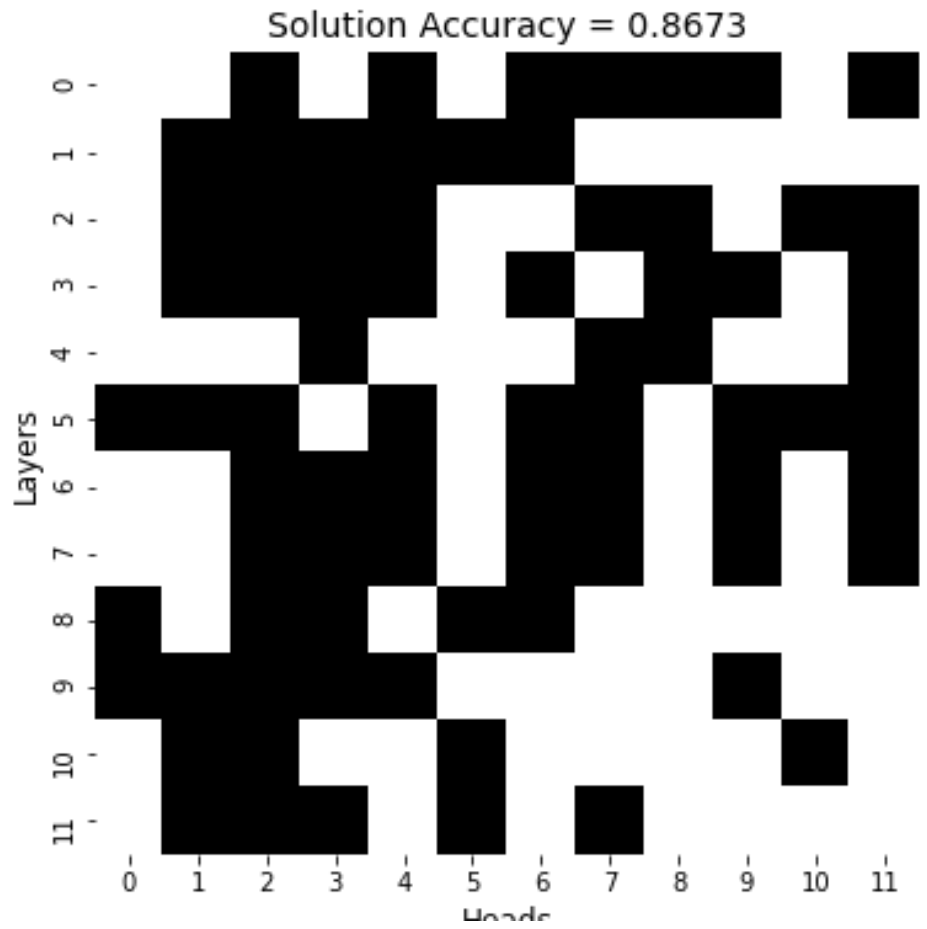

**d) Final Solution**

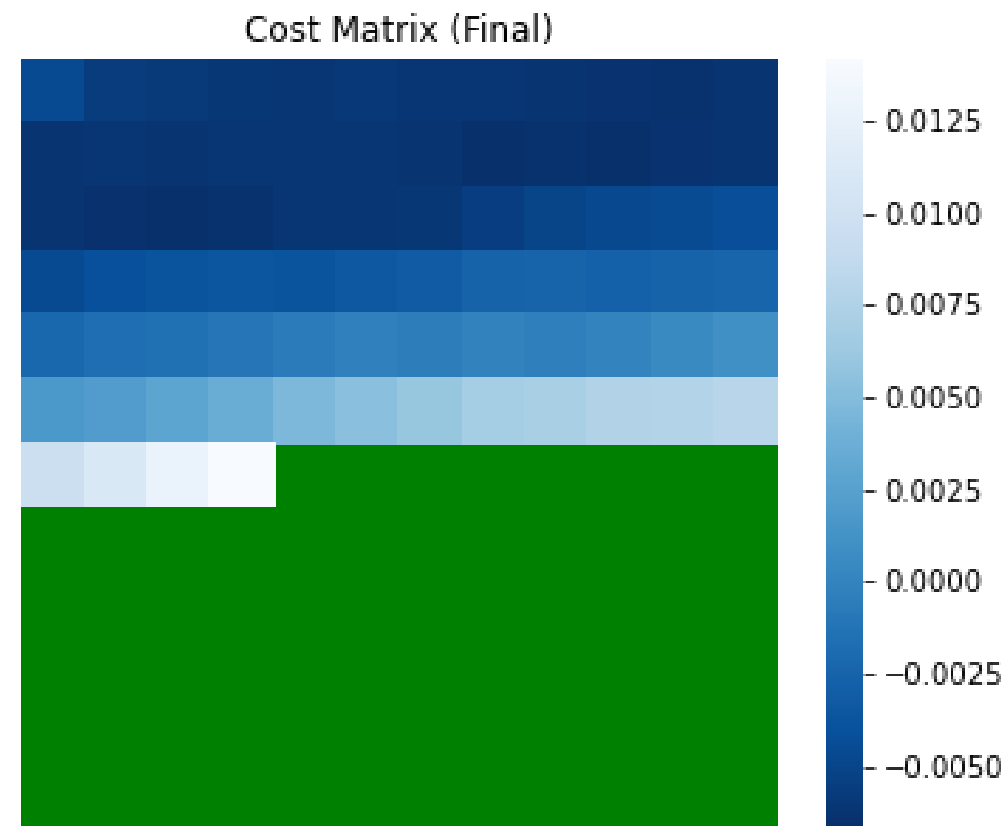

**e) Cost Matrix**

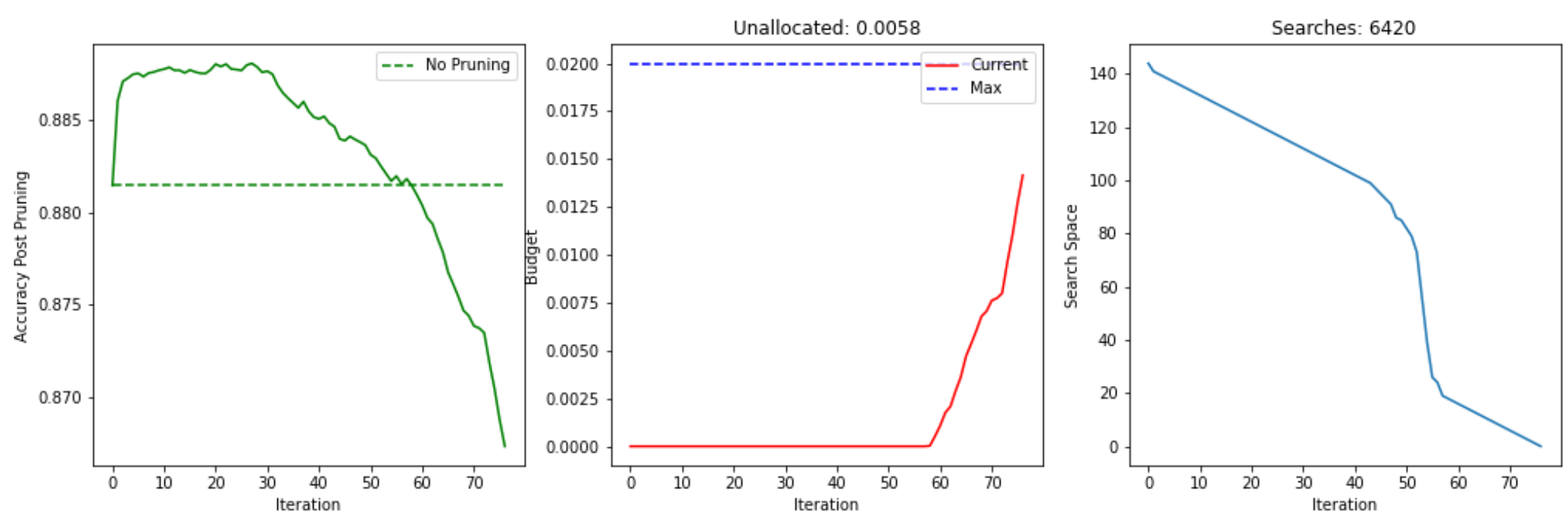

**f) Accuracy (left), Budget used (center) and # Searches (right) vs Iteration**

**Figure 15: *A** Pruning results on IMDB dataset with Budget = 2**

## 1.5 IMDB with Budget = 3

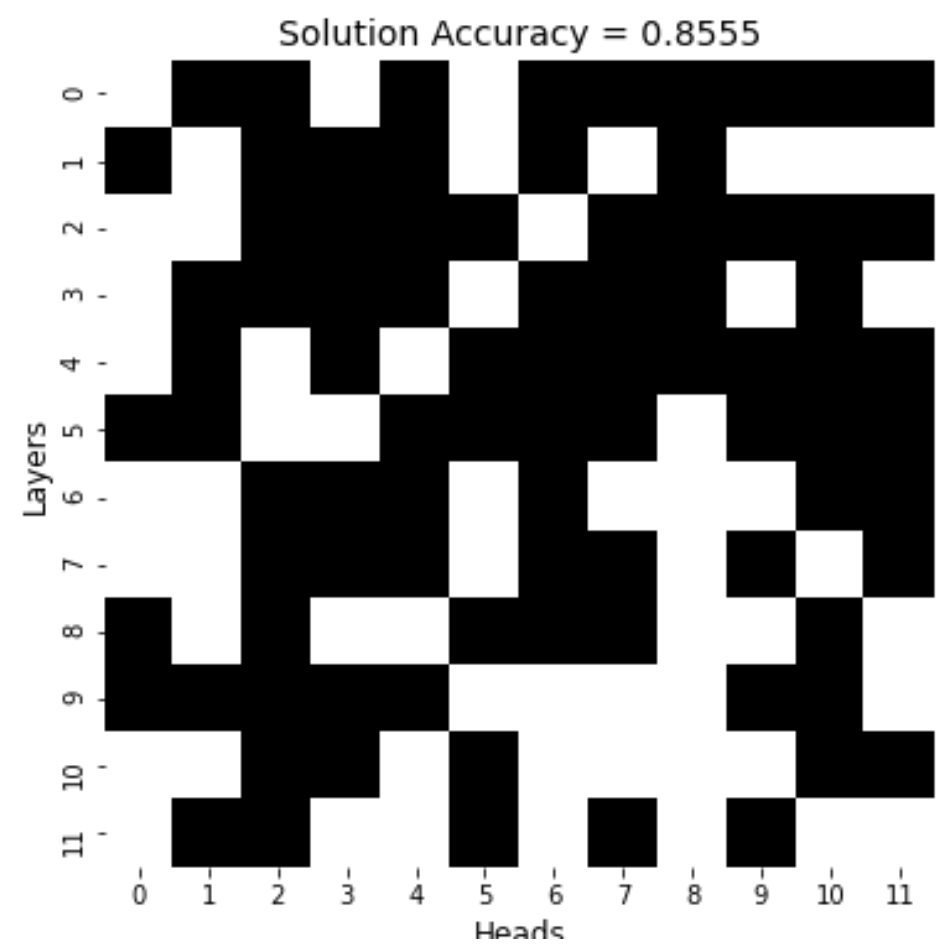

**a) Final Solution**

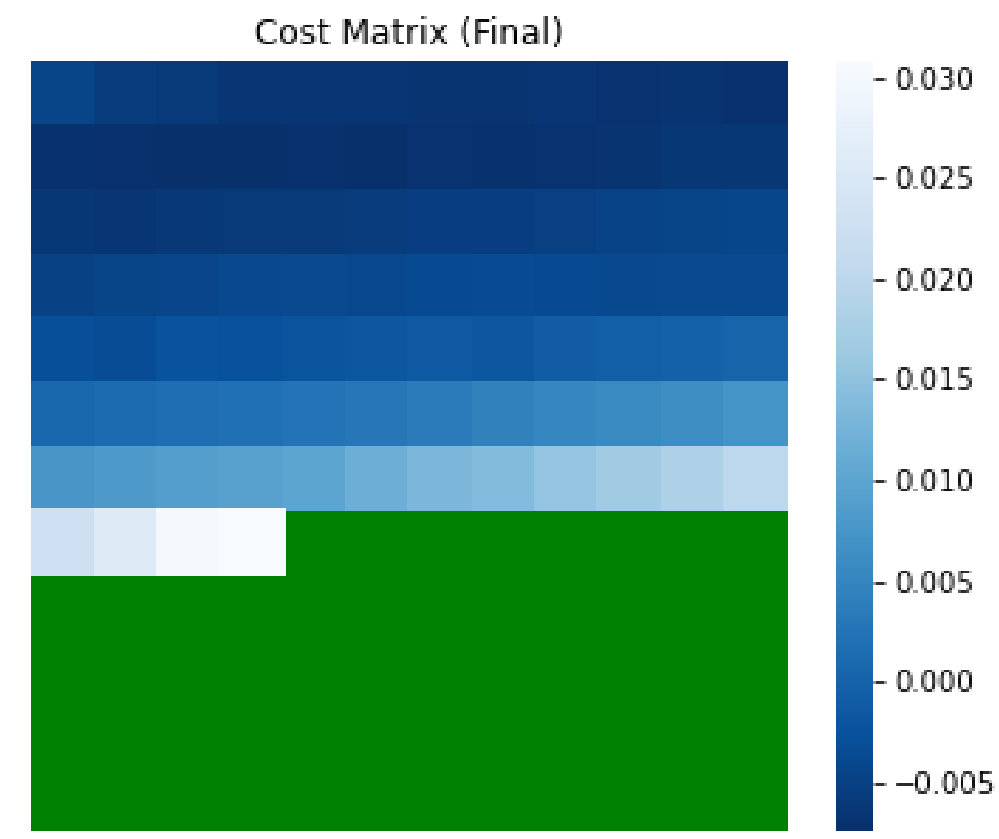

**b) Cost Matrix**

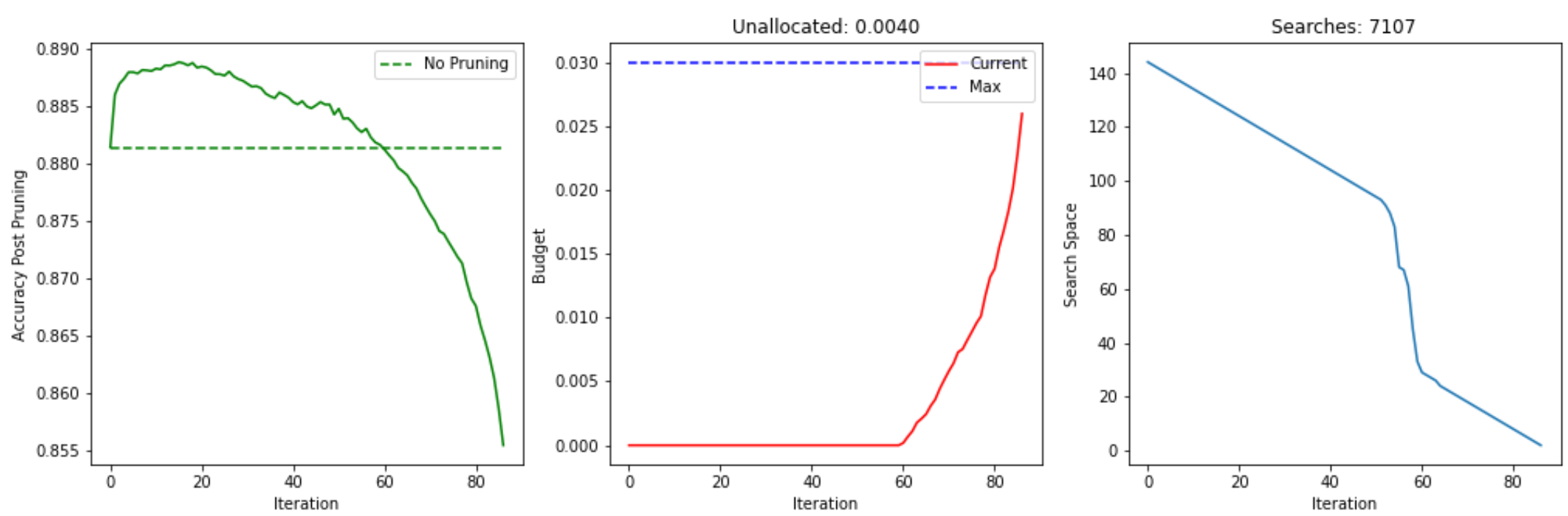

**c) Accuracy (left), Budget used (center) and # Searches (right) vs Iteration**

**Figure 16: *A** Pruning results on IMDB dataset with Budget = 3**

## 2 Random Pruning on IMDB

Figure 13 compares the result for *A* Pruning* vs Random Pruning on IMDB dataset with budget 3 and Figure 14 gives the probability of pruning X heads in a random trial.

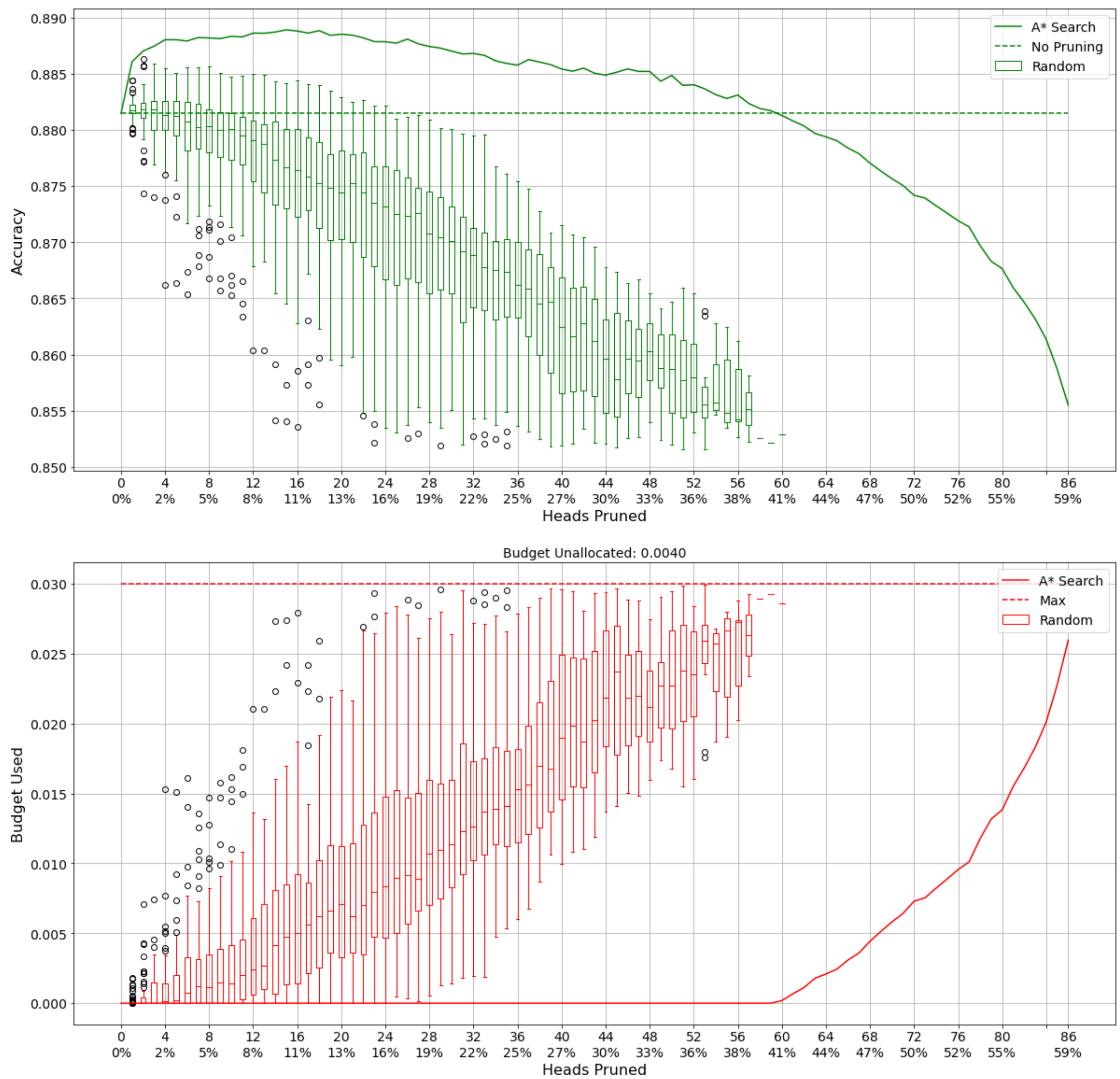

**Figure 17: Random vs *A** Pruning on IMDB dataset with budget 3.**

**Top: Accuracy vs # Heads Pruned (Compression %).**

**Bottom: Budget Used vs # Heads Pruned (Compression %)**

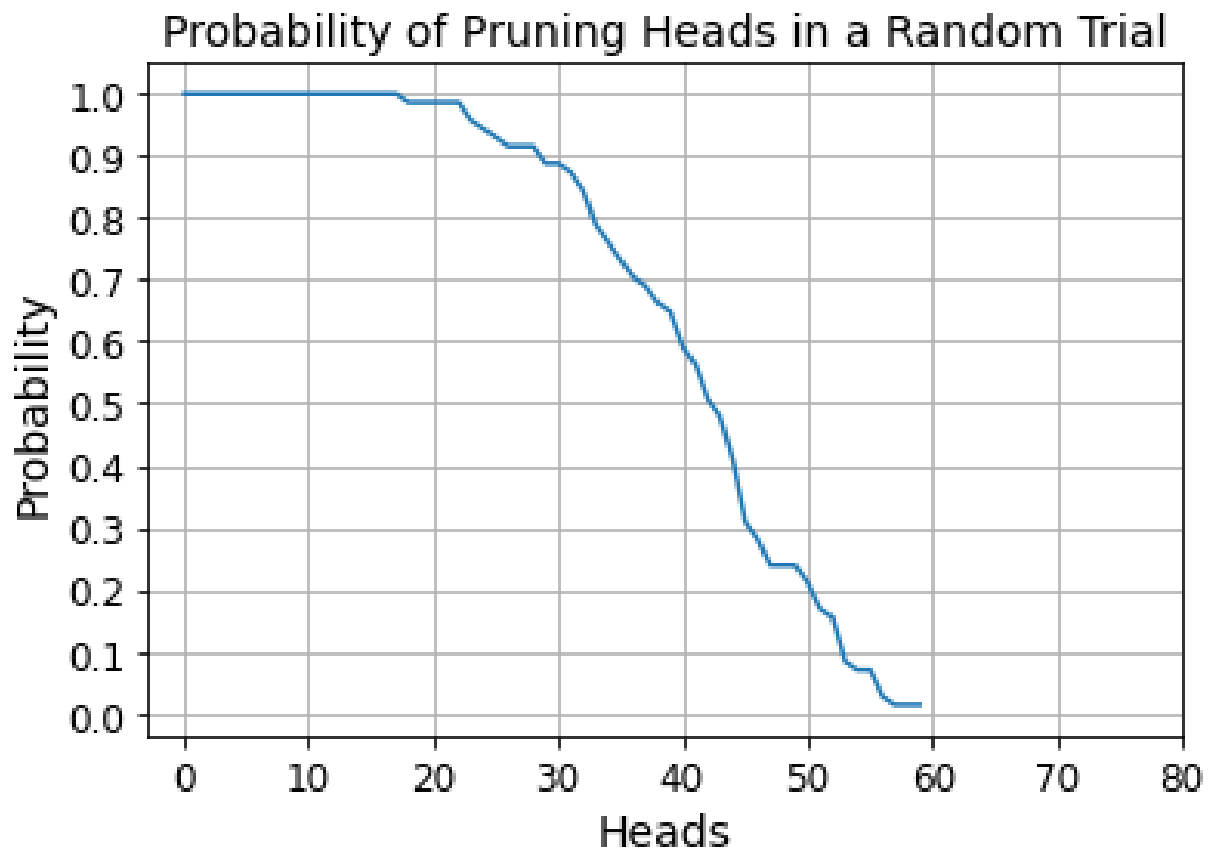

**Figure 18: Probability of pruning x heads in a random trial on IMDB dataset with budget 3**